\documentclass{article}

    \usepackage[preprint]{neurips_2024}

\usepackage{graphicx}
\usepackage[utf8]{inputenc} % allow utf-8 input
\usepackage[T1]{fontenc}    % use 8-bit T1 fonts
\usepackage{hyperref}       % hyperlinks
\usepackage{url}            % simple URL typesetting
\usepackage{booktabs}       % professional-quality tables
\usepackage{amsfonts}       % blackboard math symbols
\usepackage{nicefrac}       % compact symbols for 1/2, etc.
\usepackage{microtype}      % microtypography
\usepackage{xcolor}         % colors

\usepackage{microtype}
\usepackage{graphicx}
\usepackage{subfigure}
\usepackage{booktabs} % for professional tables
\usepackage{makecell} 
\usepackage{hyperref}

\usepackage{amsmath}
\usepackage{amssymb}
\usepackage{mathtools}
\usepackage{amsthm}

\usepackage{subcaption}
\usepackage{caption}
\usepackage{pifont}% http://ctan.org/pkg/pifont
\newcommand{\xmark}{\ding{55}}%
 
\usepackage{algpseudocode}
\usepackage{algcompatible}
\usepackage{xcolor}
\usepackage{array,multirow}
\usepackage{algorithm}
\usepackage{upgreek}
\usepackage{float}
\usepackage{newfloat}

\usepackage[colorinlistoftodos,prependcaption,textsize=tiny]{todonotes}

\newcolumntype{x}[1]{>{\centering\arraybackslash\hspace{5pt}}m{#1}}
\ExplSyntaxOn
\NewDocumentCommand{\avercalc}{O{1}+m}{%
  \clist_set:Nn \l_tmpa_clist {#2}%
  \fp_zero:N \l_tmpa_fp
  \clist_map_inline:Nn  \l_tmpa_clist {
    \fp_add:Nn \l_tmpa_fp {##1}
  }
  \fp_eval:n { round(\l_tmpa_fp/\clist_count:N \l_tmpa_clist, #1)}
}
\ExplSyntaxOff

\ExplSyntaxOn
\NewDocumentCommand{\stdcalc}{O{1}+m}{%
  \clist_set:Nn \l_tmpa_clist {#2}%
  \fp_zero:N \l_tmpa_fp
  \fp_zero:N \l_summation_fp
  \clist_map_inline:Nn  \l_tmpa_clist {
    \fp_add:Nn \l_tmpa_fp {##1}
  }
  \clist_map_inline:Nn  \l_tmpa_clist {
    \fp_add:Nn \l_summation_fp {\fp_eval:n {(##1-{\l_tmpa_fp/\clist_count:N \l_tmpa_clist})^2}}
  }
  \fp_eval:n {round(sqrt(\l_summation_fp/\clist_count:N \l_tmpa_clist), #1)}
  
}
\ExplSyntaxOff
\title{SortedNet: A Scalable and Generalized Framework for Training Modular Deep Neural Networks}

\author{
    Mojtaba Valipour\textsuperscript{\rm 1,\rm 2},
    Mehdi Rezagholizadeh\textsuperscript{\rm 2},
    Hossein Rajabzadeh\textsuperscript{\rm 1,\rm 2},\\
    \textbf{Parsa Kavehzadeh}\textsuperscript{\rm 2},
    \textbf{Marzieh Tahaei}\textsuperscript{\rm 2},
    \textbf{Boxing Chen}\textsuperscript{\rm 2},
    \textbf{Ali Ghodsi}\textsuperscript{\rm 1} \\
    \textsuperscript{\rm 1}University of Waterloo, 
    \textsuperscript{\rm 2}Huawei Noah's Ark Lab\\
    \{mojtaba.valipour, hossein.rajabzadeh, ali.ghodsi\}@uwaterloo.ca, \\
    \{mehdi.rezagholizadeh, marzieh.tahaei, boxing.chen\}@huawei.com
}
\begin{document}

\maketitle

\begin{abstract}
  Deep neural networks (DNNs) must cater to a variety of users with different performance needs and budgets, leading to the costly practice of training, storing, and maintaining numerous user/task specific models. There are solutions in the literature to deal with single dynamic or many-in-one models instead of many individual networks; however, they suffer from significant drop of performance, lack of generalization across different model architectures or different dimensions (e.g. depth, width, attention blocks), heavy model search requirements during training, and training a limited number of sub-models. To address these limitations, we propose SortedNet, a generalized and scalable training solution to harness the inherent modularity of DNNs. Thanks to a generalized nested architecture (which we refer as \textit{sorted} architecture in this paper) with shared parameters and its novel update scheme combining random sub-model sampling and a new gradient accumulation mechanism, SortedNet enables the training of sub-models simultaneously along with the training of the main model (without any significant extra training or inference overhead), simplifies dynamic model selection, customizes deployment during inference, and reduces the model storage requirement significantly. The versatility and scalability of SortedNet are validated through various architectures and tasks including LLaMA, BERT, RoBERTa (NLP tasks), ResNet and MobileNet (image classification) demonstrating its superiority over existing dynamic training methods. For example, we introduce a novel adaptive self-speculative approach based on sorted-training to accelerate large language models decoding. Moreover, SortedNet is able to train up to 160 sub-models at once, achieving at least 96\% of the original model's performance.
\end{abstract}

\section{Introduction}

% \textit{"For every minute spent organizing, an hour is earned." - Benjamin Franklin.}

\begin{figure*}[htb!]
    \centering
    \includegraphics[width=0.8\textwidth]{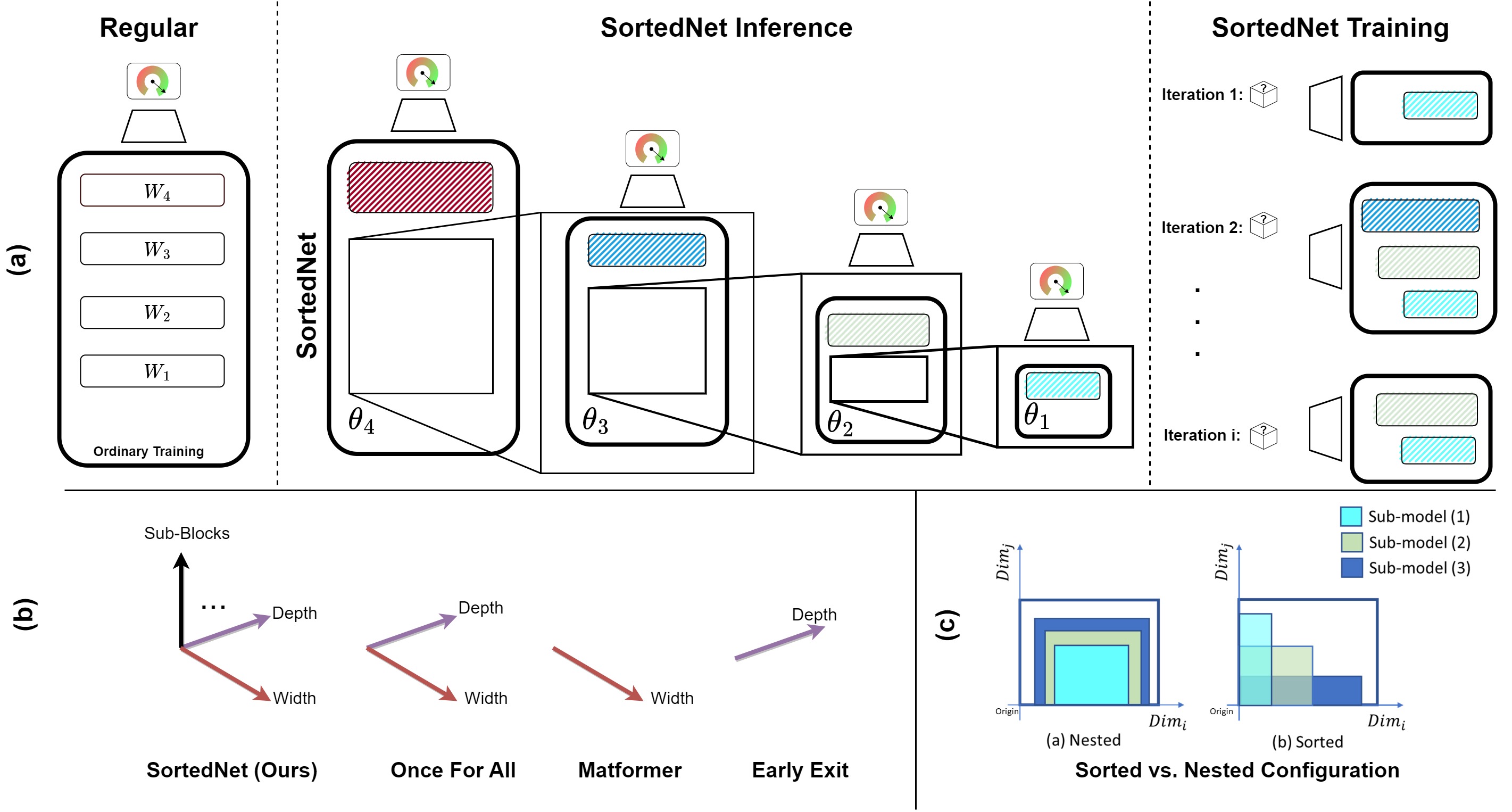}
    % \vspace{-1.35cm}
    \caption{(a) The overall diagram of our SortedNet training approach. First, we need to define the pool of sub-models of interest including the main model as well. During training, at each iteration, we sample from the pool of sub-models (given a pre-defined random distribution) to be trained for the target loss function (for one step). (b) The generalizability of Sorted configuration across more complex dimensions, supporting blocks and functions iin addition to width and depth. (c) Illustrating the difference between the nested and sorted sub-models. In nested architectures, smaller sub-models are encapsulated by larger sub-models, which is not necessarily the case for what we refer to as sorted models. Moreover, sorted models are tied to the origin (i.e. starting index) of each dimension which might not be the case in nested models.}
    \label{fig:DyLoRA} 
    % \todo[inline]{3D direction visualization figure discussed with Mehdi, comparison of SortedNet, Nested, Early Exit and Vanilla}
\end{figure*}

Deep neural networks (DNNs) are increasingly gaining interest and becoming more popular~\cite{sarker2021deep}. This popularity translates to the increasing demand and requirements from the users which should be met by these models. 
Users pre-train or fine-tune more models with various sizes to address the performance and computational needs of their tasks and target devices (with different memory and computational power  whether deployed in the cloud or on edge devices).  
However, developing, storing, maintaining, and deploying many individual models for diverse set of users can be very difficult and costly~\cite{devvrit2023matformer}. Moreover, in the era of gigantic pre-trained models~\cite{devlin2018bert, liu2019roberta} and large language models~\cite{brown2020language,chowdhery2022palm} the computational demands can vary significantly from task to task. Therefore, there is a growing demand for models which can adapt themselves to the dynamic conditions, while conventional neural network would fail to address such cases~\cite{xin2020deebert, yu2018slimmable}.

On the other hand, DNNs demonstrate modular architectures along various dimensions, like layers and blocks across depth, and neurons, channels and attention heads along width. This inherent modularity enables the extraction of sub-models with similar shapes to the original model. However, this modularity has not been deployed in regular training methods, and consequently, the performance of the sub-models falls short compared to the main model. Hence, the challenge lies in harnessing the full potential of modularity in deep neural networks, allowing for the efficient utilization of sub-models to enhance their performance and enable their practical deployment in real-world scenarios.

Instead of training individual models, we can leverage sub-models of DNNs and train them together with the main models to obtain many-in-one networks with sub-models that can be used for different tasks.  
There are variety of approaches in the literature for training sub-models~\cite{cai_once-for-all_2020, xin2020deebert, hou2020dynabert}. These techniques while effective have certain shortcomings: often use a sophisticated training process combined with knowledge distillation (which needs to train a separate teacher model) \cite{hou2020dynabert}, require architecture modification \cite{Nunez_2023_WACV}, work for specific architectures only~\cite{cai2019once}, cannot handle more than a very small number of sub-models, need heavy model search (e.g. neural architecture search) during training or inference~\cite{cai_once-for-all_2020}, involve redundant sub-model optimization \cite{fan2019reducing}, or show poor performance for the main model or sub-models~\cite{xin2020deebert}. 

To address these problems, we propose SortedNet, a generalized and scalable training solution to harness the inherent modularity of DNNs across various dimensions. As the name of our method implies, it chooses the sub-models in a sorted manner (a generalized version of nested architectures) within the main model to avoid heavy search during or after training. 
In contrast to nested models in which smaller sub-models are always totally encapsulated by larger sub-models, our generalized sorted version relaxes the nested constraint but ties the origin of sub-models to the origin of the main model across any target dimension (for more details see section ~\ref{ref::sorted}).

This sorted configuration with shared parameters enforces regular order and consistency in the knowledge learned by sub-models. One option to sort the sub-models is based on their computation and accuracy requirements which will enable us to extract our desired sub-models without requiring extensive search at the test time. 
The use of a predefined sorting order ensures that each targeted sub-model possesses a unique computation overhead, effectively removing optimization of redundant sub-models from training.

To train the sorted sub-models, we propose a novel updating scheme that combines random sampling of sub-models with gradient accumulation. We tried the SortedNet solution successfully on various architectures and tasks such as the decoder-based LLaMA (13B) large language models~\cite{touvron2023llama} on the GSM8K~\cite{cobbe2021training} mathematical reasoning task, encoder-based BERT~\cite{devlin2018bert} and RoBERTa~\cite{liu2019roberta} on the set of GLUE~\cite{wang2018glue} language understanding tasks, ResNet~\cite{he2015deep} and MobileNet~\cite{sandler2018mobilenetv2} on the CIFAR-10 image classification task. Our comprehensive empirical studies across different architectures, tasks and dynamicity along various dimensions ranging from width and depth to attention head and embedding layer show the superiority and generalizabilty of our proposed method over state of the art dynamic training methods. Moreover, SortedNet offers several benefits, including minimal storage requirements and dynamic inference capability (i.e. switching between various computation budgets) during inference. 

To summarize, the main contributions of this paper are:
\begin{itemize}
\item Introducing a many-in-one solution to configure sub-models in a sorted manner and training them simultaneously with some unique aspects such as scalability (training many sub-models), generality (CNN, Transformers, depth, width), and search-free (no need for search during training or inference among sub-models) and maintaining competitive performance for the main model. 
\item To the best of our knowledge, this work is the first attempt towards training many-in-one models along various dimensions simultaneously (and not even at different stages).
% \vspace{-5pt}
% \item We introduce a novel gradient accumulation scheme in our Sorted training based on averaging the gradients of sub-models with different dimensions instead of averaging the gradients over data batches.
\vspace{-5pt}
\item We deploy our sorted training in accelerating the decoding of large language models using a self speculative approach which can lead to about 2x inference acceleration for LLaMA 13B. 
\vspace{-5pt}
% \vspace{-5pt}
\item Demonstrating theoretical justifications and empirical evidence for the effectiveness of the proposed method. Outperforming state-of-the-art methods in dynamic training on CIFAR10 \cite{krizhevsky2009learning} with scaling the number of sub-models to 160 and achieving
at least $96\%$ of the original model’s performance.
Moreover, showing successful results on dynamic training of the BERT, RoBERTa and LLaMA models. %\todo{need more check}
\end{itemize}

\section{Related Work}

\begin{table*}[h]
\caption{Comprehensive Comparison of different existing related work and distinguishing our solution}
\centering
\resizebox{\textwidth}{!}{  
\begin{tabular}{|m{3. cm}| x{3cm} c c x{2cm}   x{3cm} c x{3.2cm}| }
\hline
\textbf{Method}& \textbf{Sub-Models Config} &  \textbf{Performance} & \textbf{Anytime}  & \textbf{Search-Free} & \textbf{\# of Sub-Models} &  \textbf{Target Dim.} & \textbf{Model}\\
 % &  &  &  &  &  &   &  &  \\
\hline
\hline\vspace{2pt}
Early Exit \cite{xin2020deebert} & Sorted  & Low & \checkmark&\checkmark & Few  & Depth & Transformer \\ 
\hline \vspace{2pt}
Layer Drop \cite{fan2019reducing} & Random  & Low &\xmark &\xmark & Many  & Depth& Transformer \\ 
 \hline \vspace{2pt}
DynaBERT \cite{hou2020dynabert} & Sorted  & High& \xmark &\xmark & Few  & Depth \& Width & Transformer  \\ 
\hline \vspace{2pt}
Once for All \cite{cai2019once}& Nested  & High & \xmark &\xmark & Many  & Width \& Depth \& Kernel & CNN \\ 
\hline \vspace{2pt}
LCS \cite{Nunez_2023_WACV} & Arbitrary  & High &  \checkmark& \checkmark & Few  & Width & CNN \\ 
\hline \vspace{2pt}
Slimmable ~\cite{yu2018slimmable} & Sorted  & Moderate & \checkmark &\checkmark & Few &   Width & CNN\\ 
\hline \vspace{2pt}
{MatFormer~\cite{devvrit2023matformer}} & Sorted  & High & \xmark & \checkmark & Few &Width &  Transformer\\
\hline \vspace{2pt}
\textbf{SortedNet (Ours)} & Sorted  & \textbf{High} & \textbf{\checkmark} & \textbf{\checkmark} & Many  &\textbf{General} & CNN \& Transformer\\ 
\hline 
\end{tabular}}

\label{tab:RelatedWork}
\end{table*}

In this section, we briefly review the most relevant existing works to our SortedNet idea. 
% We reviewed the most relevant works in the literature to our SortedNet idea.
A summary of these solutions and how they are different from each other can be found in Table~\ref{tab:RelatedWork}. For more details, please refer to appendix \ref{ref::related}.

\paragraph{ Slimmable Networks~\cite{yu_slimmable_2018}}
Slimmable networks is a width adjustable training method. It was proposed particularly for CNN architectures and thus, careful consideration of the batch normalization module for various width sizes is necessary. 
% In this regard, in slimmable networks, switchable batch normalization was used which lead to additional trainable parameters. 
In contrast to slimmable networks, our SortedNet covers more architectures and works in both depth and width dimensions. 
% \moji{It's good to note that Slimmable Networks is the summation loss!}
% \vspace{-10pt}
\paragraph{Early Exit~\cite{xin2020deebert}} is one of the most popular baseline techniques which adds a classifier to intermediate layers of an already trained model. The parameters of the main model are frozen and the classifiers are updated in a separate fine-tuning process. 
% In this approach, each of the classifiers and their subsequent network can be treated as an independent sub-model. 
While this solution is relatively straightforward, the performance of the sub-models lags significantly behind that of the main model. 
% Also dedicating a separate classification head to each sub-model can significantly increase the memory demand at inference. 
% \vspace{-10pt}
 \paragraph{Dayna-BERT~\cite{hou2020dynabert}}

presents a dynamic compression method for pre-trained BERT models, enabling flexible adjustments in model size, both in depth and width, during inference. DynaBERT is different from us in the follow aspects: first, in DynaBERT, only a very few sub-models are functional; second, DynaBERT requires an already trained teacher model and utilizes knowledge distillation (KD); third, DynaBERT needs search to find an optimal sub-model; last, DynaBERT is architecture dependent.
% \vspace{-10pt}
\paragraph{Layer-drop~\cite{fan2019reducing}} 
is a structured dropout training which allows layer pruning at the inference time. Similar to DynaBERT, it is applied to pre-trained language models; however, in contrast to DynaBERT, Layer-drop only targets the depth of neural networks and not their width. 

% \vspace{-10pt}
\paragraph{Once-for-All (OFA)~\cite{cai_once-for-all_2020}} targets efficient inference across different devices. It first trains a network which supports many sub-models with varying latency/accuracy characteristics ; it then searches among the feasible sub-models according to the accuracy and latency requirements of their target device. 
OFA is different from our solution in: first, it has a progressive training nature in contrast to our stochastic or summation loss;
second, it needs teacher and KD; third, it requires a separate neural architecture search (NAS) at the inference time; fourth, OFA is for CNN-based models; last, it does not have any particular assumption for configuring sub-models (see Figure~\ref{fig:comp} for more details).   

% \vspace{-10pt}
\paragraph{Learning Compressible Subspace (LCS)~\cite{Nunez_2023_WACV}} is an adaptive compression technique based on training compressible subspace of neural networks. While LCS does not require any re-training at the inference time,  this solution has some other limitations including: first, it needs double memory at the training time; second, the choices of initial weights and the compression function are unclear and arbitrary (left as a hyper-parameter); third, it is only tried on CNNs; forth, similar to Layer-drop, the search space of sub-models is huge which makes the training sub-optimal.
% \vspace{-10pt}
\paragraph{MatFormer~\cite{devvrit2023matformer}} is a pre-training only many-in-one solution based on summation loss for Transformer-based models. MatFormer works only along the width dimension of the FFN block in Transformers and cannot handle more than a very few number of sub-models.

\section{Proposed Method}
\label{sec:method}

\subsection{A Generalized and Scalable View}
In the related work section, we have discussed several approaches concerning the training of many-in-one networks. These approaches differ in terms of their target architecture, training loss, number of training parameters, the configuration of the sub-models (random, nested, or sorted), the number of trained sub-models, and reliance on search or re-training before deployment. 
Our SortedNet method can be viewed as a simple, general, and scalable version of these existing solutions. These benefits have mostly resulted from the sorted configuration of sub-models with their shared parameters and our stochastic training. 
% To the best of our knowledge, this is the first work that has scaled training of sorted sub-models to various dimensions and different architecture types.
% In this subsection, we present SortedNet, a generalized and scalable method for training many-in-one neural networks. 
In order to train many-in-one networks, we need to specify a few design choices: first, how to form the sub-models and their configurations; second, what are the target architectures; and third, how to train the sub-models along with the main model.

\paragraph{Designing the sub-models}
SortedNet imposes an inductive bias on training based on the assumption that the parameters of sub-models have a concentric architecture tied to the origin along each dimension (which we refer to as a \textit{sorted architecture}).
This sorted configuration with shared parameters enforces a regular order and consistency in the knowledge learned by sub-models (see Figure.~\ref{fig:DyLoRA}).

\paragraph{Sorted vs. Nested Architectures}
\label{ref::sorted}
In this work, we introduce the term of \textit{sorted} architectures to extend and generalize the concept of nested architectures. 
In contrast to nested models in which smaller sub-models are always totally encapsulated by larger sub-models, our sorted sub-models would be tied to the origin (starting index) of each dimension independently. 

% \begin{figure}[htb!]
% \centering
% \resizebox{0.65\columnwidth}{!}{  
% \includegraphics[width=1.2\textwidth]{Figs/Fig_appndix.png}
% }
% \vspace{-15pt}
% \caption{Illustrating the difference between the nested and sorted sub-models. In nested architectures, smaller sub-models are encapsulated by larger sub-models, which is not necessarily the case for what we refer to as sorted models. Moreover, sorted models are tied to the origin (i.e. starting index) of each dimension which might not be the case in nested models.}
% % \label{fig:appndix}
% \end{figure}
 
Let's consider a many-in-one neural network $f(x;\theta (n))$ with the parameters $\theta(n)$ and the input $x$ which is comprised of $n$ sub-models $f(x;\theta (i)) |_{i=0}^{n-1}$, where $\theta(i)$ represents the weights of the $i^{\text{th}}$ sub-model. We define a universal set which contains all unique sub-models: $\Theta= \{\theta(0), \theta(1), ..., \theta(n) \}$.  

\paragraph{Setting up an order} Suppose that we would like to target $D = \{ Dim_1, Dim_2, ..., Dim_K\}$ many-in-one dimensions in our model. Then, let's start with $\Theta = \varnothing$ and build the sub-models iteratively. In this regard, at each iteration $t$ during training, we have sampling and truncation procedures along any of the targeted dimensions: 
\begin{equation}
\begin{split}
        & \theta_t^* = \cap_{j=1}^{|D|} \theta_{Dim_j\downarrow b_j^t}(n)  \text{  ~~~ where } b_j^t \sim P_{B_j} \\ 
        &\text{If } \theta_t^*  \notin \Theta \text{ : } \Theta \leftarrow \Theta \cup \{ \theta_t^* \} 
\end{split}
\label{eq:1}
\end{equation}
where $Dim_j\downarrow b_j^t$ indicates that we have truncated $\theta(n)$ along the $Dim_{j}$ dimension from the index 1 up to the index $b_j^t$ at the iteration $t$. $b_j^t$ is sampled from a distribution $P_{B_j}$ with the support set of $B_j = \{1,2, ..., d_j\}$ to form the $i^{\text{th}}$ sub-model. $d_j$ refers to the maximum index of the $j^{\text{th}}$ dimension. This iterative process will be done during training and the set of $n$ unique sub-models $\Theta$ will be built.

To illustrate the process better, let's see a simple case such as $\text{BERT}_{base}$ where we want to make a many-in-one network across the width and depth dimensions, $D = \{ \text{Depth}, \text{Width} \}$. In this case, we have 12 layers and a hidden dimension size of 768. Suppose that $Depth$ corresponds to $j=1$ and $Width$ corresponds to $j=2$ in Eq.~\ref{eq:1}. For simplicity, let's use a discrete uniform distribution for sampling indices across these two dimensions. To create the first sub-model ($i=1$), we need to sample $b_1^1$ uniformly from the set of natural numbers in the range of 1 to 12: $B_1 = \{1,2,..., 12\}$; and we need to sample $b_2^1$ from the range of 1 to 768: $B_2 = \{1,2,3,..., 768\}$. Bear in mind that we can even choose a subset of $B_1$ and $B_2$ as the support set for sampling probability distribution. After these two samplings, we will have two truncated sets of parameters: $\theta_{Depth \downarrow b_1^1 }$ and $\theta_{Width \downarrow b_2^1 }$. The intersection of these two truncated parameters will give us the first sub-model:  
$\theta_1 = \theta_{Depth \downarrow b_1^1} \cap \theta_{Width \downarrow b_2^1} $.   

\paragraph{Training Paradigm}
Regular training of neural networks concerns improving the performance of the whole model and usually this training is not aware of the performance of the sub-models. In fact, in this scenario, if we extract and deploy the sub-models of the trained large model on a target task, we would experience a significant drop in the performance of these sub-models compared with the main model. 
However in SortedNet, we propose a training method that allows for training sub-models together with the main model in a stochastic way. 
The SortedNet paradigm leads to the following benefits: 
\vspace{-5pt}
\begin{itemize}
    % \item The cost of obtaining sub-models is equal to training the large model only once. This leads to significant reduction to SOTA works that require separate training for each sub-model. 
    % \item Minimize the performance drop of the large model compared to normal training. 
    % \vspace{-5pt}
    \item Search-free sub-model extraction: after training, by importance sorting of sub-models the best sub-model for a given budget can be selected without the need for search. 
    % \vspace{-10pt}
    \item Anytime: Each smaller sub-model is a subset of a larger one which makes switching between different sub-models efficient.
    This leads to an important feature of our SortedNet which is so-called \textit{anytime} that is a network which can produce its output at any stage of its computation. 
    % \vspace{-10pt}
    \item Memory efficient:  we train a many-in-one network where sub-models are all part of a single checkpoint, which minimizes storage requirement.
\end{itemize}
\vspace{-5pt}
For efficiency purposes, in our training, the last layer, e.g. the classification layer, is shared between all sub-models; alternatively, we can add a separate classification layer to each sub-model. For simplicity and efficiency, we chose to the former i.e. use a shared classification layer.

% \begin{figure}[h!]
% \hspace*{-1cm}
% \centering
% \resizebox{0.7\columnwidth}{!}{  
% %\input{cifar10-160.tex}
% \includegraphics[]{Figs/cifar10-classification-acc-figure.png}
% }
% \caption{CIFAR10 classification accuracy (and recovery percentage) for Sorted-Net (160 Models) and the baseline. In each cell, we reported the performance of the sub-model (top) and the relative performance of the model (in percentage) with respect to the baseline largest model performance (bottom). W. Only: Sorting only the widths, D. Only: Sorting only the depth. More black the better.}
% \label{fig:scalable160}
% \end{figure}

\begin{figure}[htb!]
    \centering
    \includegraphics[width=\textwidth]{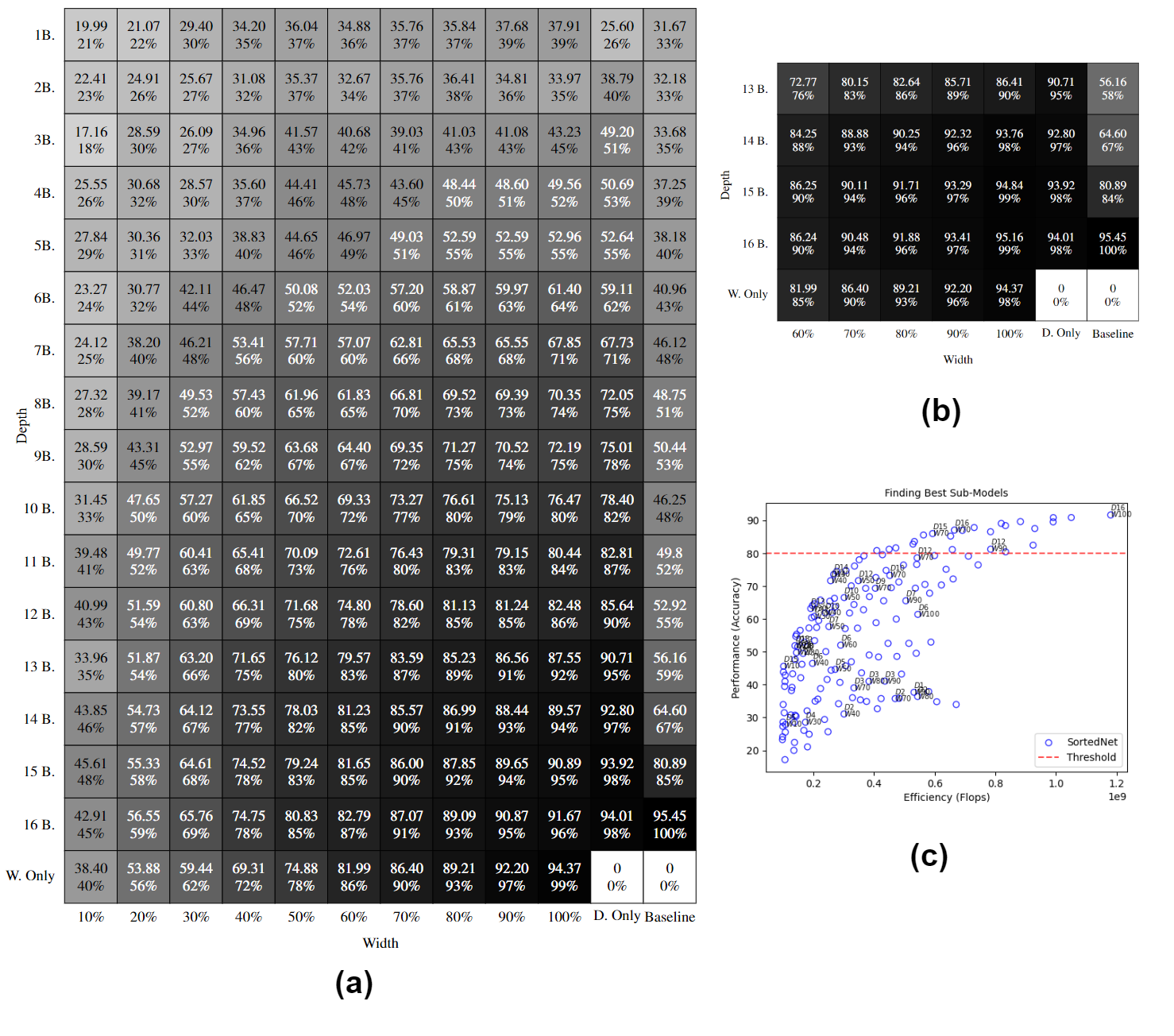}
    % \begin{subfigure}%[t]
    %     \centering
    %     \includegraphics[height=1.9in]{Figs/cifar10-classification-acc-best-performing-subsets-figure.png}
    %     % \subcaption{(a)}
    %     % \label{fig:bestsubset}
    % \end{subfigure}%
    % \hfill
    % \begin{subfigure}%[t]
    %     \centering
    %     \includegraphics[height=2.08in]{Figs/plot-Elbow.png}
    %     % \label{fig:elbow}
    %     % \subcaption{(b)}
    % \end{subfigure}
    \vspace{-4pt}
    \caption{(a) CIFAR10 classification accuracy (and recovery percentage) for Sorted-Net (160 Models) and the baseline. In each cell, we reported the performance of the sub-model (top) and the relative performance of the model (in percentage) with respect to the baseline largest model performance (bottom). W. Only: Sorting only the widths, D. Only: Sorting only the depth. More black the better. (b) CIFAR10 classification performance for the best-performing subset of sub-models trained by SortedNet from scratch. More black the better. (c) Finding best sub-models automatically using a desired threshold bar to eliminate the worst performing models.}
    \label{fig:scalable160-bestsubset-left-elbow-right}
\end{figure}

\subsection{SortedNet Algorithm}

In this subsection, we describe our proposed training algorithm. For training a SortedNet with $n$ sub-models, at each iteration during training, a random index needs to be sampled from a pre-defined distribution: $b_j^i \sim P_{B_j}$. 
% where $1 \leq b_i \leq n$ and $\theta_0 \subseteq \theta_1 \subseteq ... \subseteq \theta_n$. 
After finding the target sub-model $\theta_t^*$ at each iteration, we can use one of the following objectives to update the parameters of the selected sub-model: 

\begin{itemize}
    \item (Stochastic Loss) Only train the selected sub-model $f(x,\theta_t^*)$ : \\$\underset{\theta_t^*}{\min} ~\mathcal{L} \triangleq \text{L}(y, f(x,\theta_t^*))$
    where $L$ is the loss function for training the model on a given task (e.g. $L$ can be a regular cross entropy loss) and $y$ refers to the ground-truth labels.
    \item (Stochastic Summation) Train the sub-model $f(x,\theta_t^*)$ and all its targeted sub-models along each dimension. Let's assume that $\Theta^\perp (\theta_t^*)$ is the universal set for all targeted sub-models of $\theta_t^*$. Then the loss function can be defined as:  \\
    $\underset{\Theta^\perp (\theta_t^*)}{\min} ~\mathcal{L} \triangleq \sum_{\theta \in \Theta^\perp (\theta_t^*)} \text{L}(y, f(x,\theta) )$
    % \item Train the sub-models $SM_{1,w_i}$ to $SM_{L,w_i}$: \\
    % $\underset{\bigcup_{i=1}^{L} \theta_i}{\min} ~\mathcal{L} \triangleq \sum_{i=1}^{L} \text{CE}(y, SM_{i,w_i}(x;\theta_i) )$. 
\end{itemize}

This way, one sub-model or a subset of sub-models are updated in each iteration. Alternatively, one can choose to train all the sub-models at each iteration which is costly in the large scale. %(see Fig. \ref{fig:method})

\subsection{Why Does SortedNet Work? }
In Appendix \ref{ap:theory}, we provide theoretical justification for parameter convergence of the sub-models in identically trained scenarios and also provide the performance bound between the trained sub-models and their similar corresponding network trained independently. 
\vspace{-10pt}
\paragraph{Convergence} Suppose $\hat{f}$ is a sub-model of a larger network, and $f$ is an identical model architecture trained independently. We aim to understand the relationship between the parameters of these two networks, $\theta$ for $\hat{f}$ and $\phi$ for $f$, as they are trained under identical conditions.
Assuming that the gradients of the loss functions for $\hat{f}$ and $f$, 
% denoted as $\mathcal{L}_{\hat{f}}$ and $\mathcal{L}_f$ respectively, 
are $L$-Lipschitz continuous, and the learning rate is $\eta$, we show that 
\begin{equation}
    \|\theta_{t+1} - \phi_{t+1}\| \leq (1 + \eta L) \|\theta_t - \phi_t\|.
\end{equation}
This indicates that the difference in the parameters of $\hat{f}$ and $f$ is governed by the Lipschitz constant $L$ and the learning rate $\eta$, suggesting that the parameters should remain close throughout the training process, especially when the difference between the gradients of the loss functions of the two networks is negligible.
\vspace{-10pt}
\paragraph{Performance Bound} Moreover, we would like to find a performance bound between a trained sub-model (with optimized parameters $\theta^*$) and its corresponding individual model (with optimized parameters $\phi^*$). 
% In this regard, supposed that we have a sub-model $\hat{f}(x;{\theta}^*)$ and its corresponding individual model ${f}(x;{\phi}^*)$. 
Let's assume that ${\phi}^* = {\theta^*} + \Delta \theta$.
We show in Appendix~\ref{ap:theory} that the deviation $ \Delta {f} = {f}(x;{\phi^*}) - \hat{f}(x;{\theta^*}) $ in the function value from its optimal value due to a parameter perturbation \( \Delta \theta \) is bounded by \( \frac{1}{2} L \|\Delta \theta\|^2 \) under the assumption of L-Lipschitz continuity of the gradient. 

\begin{equation}
\begin{split}
    &\Delta {f} \approx \frac{1}{2} \Delta \theta^T H(x; \theta^*) \Delta \theta \leq \frac{1}{2} L \|\Delta \theta\|^2
\end{split}    
\end{equation}

This result implies that the function value's deviation grows at most quadratically with the size of the parameter perturbation.

% \subsection{SVD Interpretation}
% \todo[inline]{explain the svd in uni-dimension setting and then expand why svd is limited to tensors and not functions}

% In Theorem 2 of NestedDropout \cite{rippel2014learning}, it has been proved that every optimal solution of the nested dropout problem must be of the form . Under the orthonormality constraint, it has been shown that the optimal solution for each b-truncation problem recovers exactly the set of the K top eignevectors, ordered by eigenvalue magnitude. 

% \subsection{Why better than Early Exit?}
% \todo[inline]{explain why SortedNet is better than Early Exit}

% \moji{SortedNet mechanism is better than Early Exit simply because it uses the non-linear capacity of sub-blocks.}  

\section{Experiments}

In this section, we conduct a set of experiments to show the effectiveness and importance of our solution. The details of the hyper-parameters for each experiment can be found in Appendix \ref{ap:hyperparameters}.

\subsection{Is SortedNet Scalable?}

To show that our proposed method is scalable, we designed an experiment that try to train 160 different models across multiple dimensions (width and depth) all at once. As baseline, we trained the largest network (a MobileNetV2), and reported the best performance of the model. Because the performance of the model was poor for all the other sub-models (less than 12\%), we trained the classifier layer for 5 more epochs before evaluating each sub-model for the baseline and reported the best performance. As the results suggests in Figure \ref{fig:scalable160-bestsubset-left-elbow-right}-a, our method was able to capture the maximum performance for many of these sub-models in a zero-shot manner. In each cell, we reported the performance of the sub-model on top and the recovery percentage of the model with respect to the largest model (in this example, 95.45). Despite sharing the weights across all models, sharing the classifier and zero-shot evaluation, the proposed method preserved up to 96\% of the performance of the largest model which is highly encouraging. Further training of the classifier for our proposed method will lead to even better performance as shown in Appendix \ref{ap:adjusting-classifier} (between $\sim$2 to 15\% improvement for different sub-models). In addition, we also tried to sort the depth and width using proposed method individually, which is reported in the Figure \ref{fig:scalable160-bestsubset-left-elbow-right}-a as D. Only, and W. Only, respectively. Across width, SortedNet successfully preserved up to 99\% of the largest network's performance.

\subsection{Can we find the best sub-models using SortedNet?}

% \begin{figure}[htb!]
%     \centering
%     \begin{subfigure}%[t]
%         \centering
%         \includegraphics[height=1.9in]{Figs/cifar10-classification-acc-best-performing-subsets-figure.png}
%         % \subcaption{(a)}
%         % \label{fig:bestsubset}
%     \end{subfigure}%
%     \hfill
%     \begin{subfigure}%[t]
%         \centering
%         \includegraphics[height=2.08in]{Figs/plot-Elbow.png}
%         % \label{fig:elbow}
%         % \subcaption{(b)}
%     \end{subfigure}
%     \caption{(left) CIFAR10 classification performance for the best-performing subset of sub-models trained by SortedNet from scratch. More black the better. (right) Finding best sub-models automatically using a desired threshold bar to eliminate the worst performing models.}
%     \label{fig:bestsubset-left-elbow-right}
% \end{figure}

As shown in Figure \ref{fig:scalable160-bestsubset-left-elbow-right}-b, based on the performance of the models in the previous experiment shown in Figure \ref{fig:scalable160-bestsubset-left-elbow-right}-a, we selected a subset of best-performing networks (width $>$ 60\% and depth $>$ 13 blocks), and retrained the network from scratch using SortedNet to show the success rate of our proposed method. As shown, SortedNet successfully preserved up to 99\% of the performance of the ordinary training of the largest network. %Alternatively, one can continue the training for the chosen subset using SortedNet.
We can also make this selection process fully automated by sorting the performance of all sub-models after evaluation and filtering out a subset of best-performing models that perform better than a desired threshold. As it can be seen in Figure \ref{fig:scalable160-bestsubset-left-elbow-right}-c, there is a set of sub-models which perform better than 80\%. To better understand the pattern, we annotated some of the points using ``$_{W}^{D}$" as template which shows for each model the corresponding width and depth.

% \todo[inline]{customization and user customization}

\subsection{Can we generalize SortedNet?}

\begin{table*}[htb!]
\caption{Comparing the performance of state-of-
the-art methods with Sorted-Net over CIFAR10 in terms of test accuracy.}
\centering
\resizebox{\textwidth}{!}{  
\begin{tabular}{lcc|ccc|ccc}
\hline
Network & Width & FLOPs & NS-IN & LCS-p-IN  & SortedNet-IN & NS-BN  & LCS-p-BN (aka US) &SortedNet-BN\\
\hline
\multirow{4}{*}{cpreresnet20 \cite{he2015deep} (CIFAR10)} & 100\% & 301M & 88.67$\pm$2.1 & 87.61$\pm$2.3 & \textbf{89.14}$\pm$2.1 & 79.84$\pm${5.0}  & 65.87$\pm$2.1  & \textbf{85.24}$\pm${2.3}\\
& 75\% & 209M & 87.86$\pm$1.6 & 85.73$\pm$2.2 & \textbf{88.46}$\pm$2.1 & 78.59$\pm$3.4  & \textbf{85.67}$\pm$1.4 &85.29$\pm${3.2}\\
& 50\% & 97M & 84.46$\pm$2.1 & 81.54$\pm$5.3 &\textbf{85.51}$\pm$2.2 & 69.44$\pm$4.0  & 65.58$\pm$3.1  &\textbf{70.98}$\pm${4.3} \\
& 25\% & 59M &75.42$\pm$2.2 & \textbf{76.17}$\pm$1.2 &75.10$\pm$2.6 & 10.96$\pm$3.4  & \textbf{15.78}$\pm$3.5  &12.59$\pm${3.2}\\
\hline
avg. & -& - & 84.10 & 82.76 & \textbf{84.55} & 59.70 & 58.22 & \textbf{63.52}\\
\hline
\end{tabular}}

\label{tab:CIFAR10-lcs}
\end{table*}

% ref: http://www.cs.toronto.edu/~rgrosse/courses/csc321_2017/slides/lec6.pdf

% \todo{introduce baselines}
In another experiment, as shown in Table \ref{tab:CIFAR10-lcs}, we demonstrate the superiority of our stochastic approach compared to the state-of-the-art methods such as LCS (shown as $LCS_p$ in the table) \cite{Nunez_2023_WACV}, Slimmable Neural Network (NS) \cite{yu2018slimmable}, and Universally Slimmable Networks (US) \cite{yu2019universally}. To make the comparisons fair, we equalized the number of gradient updates for all models. % based on the principle/rule of thumb that a backward pass worth approximately twice the forward pass for all methods \footnote{https://www.alignmentforum.org/posts/fnjKpBoWJXcSDwhZk/what-s-the-backward-forward-flop-ratio-for-neural-networks} \cite{Kaplan2019NotesOC}. 
We also tried to remove the impact of architecture design such as the choice of the normalization layers. Therefore, we tried to compare methods by different layer normalization techniques such as BatchNorm \cite{ioffe2015batch} and InstanceNorm \cite{ulyanov2016instance}. In addition, we ensure that  complementary methods such as Knowledge Distillation have no impact the results as these methods can be applied and improve the results independent of the method. As shown in the table, SortedNet demonstrates a superior average performance compared to other methods, indicating its generalization across various settings such as different norms. It is worth noting that we realized the unexpected nature of the LCS-p-BN results in Table \ref{tab:CIFAR10-lcs}. However, these results are in line with the original LCS paper's observations \cite{Nunez_2023_WACV} (see Figure 3 of the LCS paper). The LCS authors \cite{Nunez_2023_WACV} also hypothesized that this drop caused by inaccurate batch norm statistics. To address this, they suggested an architectural adjustment to GroupNorm. Our SortedNet approach, on the other hand, remains unaffected by this issue, thus requiring no such modifications.

\subsection{Extending Sorted Net to Pre-trained Language Models}

\begin{table*}[h!]
\caption{A comparison of the performance of different sub-models with and without the SortedNet. The model's performance will improve if we have more budgets and calculate the representation of deeper layers.}
\centering
\resizebox{0.9\textwidth}{!}{  
\begin{tabular}{lccccccccc|c}
% \hline
% \multicolumn{11}{c}{\textbf{Refine the Performance Over Time}} \\
\hline
& Acc. & Acc. & F1 & Mathews Corr. & Acc. & Acc. & Acc. & Pearson & \\
\textbf{Model} & \textbf{MNLI}& \textbf{SST-2}& \textbf{MRPC}& \textbf{CoLA}& \textbf{QNLI}& \textbf{QQP}& \textbf{RTE}& \textbf{STS-B}& \textbf{AVG (Ours)}& \textbf{AVG w/o ours}\\
\hline
Sorted-RoBERTa (1L)&	60.07&	70.76&	81.22&	0.00&	59.64&	77.80&	47.65&	9.36&	\textbf{50.81}& 40.33\\
Sorted-RoBERTa (2L)&	71.98&	80.28&	81.22&	0.00&	81.49&	87.09&	47.29&	70.37&	\textbf{64.97}&40.86\\
Sorted-RoBERTa (4L)&	76.74&	80.50&	81.22&	0.00&	85.21&	88.82&	46.93&	75.07&	\textbf{66.81}&41.06\\
Sorted-RoBERTa (4L)&	79.13&	84.75&	81.22&	44.51&	86.60&	90.11&	49.10&	84.94&	\textbf{75.04}&42.95\\
Sorted-RoBERTa (5L)&	81.14&	89.91&	81.22&	48.41&	87.88&	90.86&	55.96&	88.22&	\textbf{77.95}&43.80\\
Sorted-RoBERTa (6L)&	82.21&	92.09&	86.67&	53.41&	88.83&	91.12&	67.87&	89.09&	\textbf{81.41}&46.13\\
Sorted-RoBERTa (7L)&	82.99&	92.78&	89.13&	56.42&	89.29&	\textbf{91.29}&	73.29&	89.58&	\textbf{83.10}&44.80\\
Sorted-RoBERTa (8L)&	83.33&	93.23&	89.78&	57.22&	89.40&	91.29&	75.09&	89.67&	\textbf{83.63}&55.17\\
Sorted-RoBERTa (9L)&	83.39&	92.66&	89.66&	58.69&	89.40&	91.25&	\textbf{77.26}&	89.72&	\textbf{84.00}&61.36\\
Sorted-RoBERTa (10L)&	\textbf{87.42}&	93.12&	\textbf{91.64}&	\textbf{61.21}&	\textbf{91.87}&	91.19&	74.01&	\textbf{89.74}&	\textbf{85.02}& 54.30\\
Sorted-RoBERTa (11L)&	87.34&	\textbf{93.35}&	91.45&	60.72&	91.74&	91.17&	74.01&	89.72&	\textbf{84.94}&77.48\\
Sorted-RoBERTa (12L)&	83.35&	92.89&	90.81&	59.20&	89.44&	91.28&	76.53&	89.77 & 84.16 & \textbf{86.13}\\
\hline 
avg. & 79.26 & 87.93 & 86.09 & 41.25 & 85.50 & 89.45 & 64.26 & 79.61 & \textbf{76.67} & 52.86 \\
\hline
\end{tabular}}

\label{tab:sortedRoBerta}
\end{table*}

In this experiment, the goal is to apply SortedNet for a pre-trained transformer model and evaluate the performance on the GLUE benchmark \cite{wang2018glue}. As the baseline, we chose RoBERTa \cite{liu2019roberta} to demonstrate the flexibility of our algorithm. In Table \ref{tab:sortedRoBerta}, we sorted all the layers of  RoBERTa-base. As the results demonstrate, our proposed method in average perform better than the baseline by a significant margin ($\sim$ 23\%). However, the largest model has a small drop in performance (less than 2\%). It is interesting that the transformer architecture can preserve the performance of sub-models up to some extent without additional training. However, our algorithm improve the performance of these sub-models between 10 to 40\% approximately. A more complex setting (sorting across Bert models), has been investigated in Appendix \ref{ap:extend-more-complex}.

% \todo[inline]{Adaptive self-speculative method introduction}
\begin{table*}[t]
\caption{
  Speed-up in inference time on GSM8K benchmark by utilizing Speculative Decoding and Adaptive Early-Exit Techniques over SortedNet models. 
  % \todo[inline]{Table 4-GSM8K, figure explanation}
  % \vspace{-4mm}
 }
 \setlength\extrarowheight{2pt}
 \centering
 \scalebox{0.65}{\begin{tabular}{l|ccc|ccc}
  
  % \cline{1-3}
  \multicolumn{7}{c}{\textbf{SortedNet Efficient Decoding}} \\
  \toprule
   \multicolumn{4}{c}{\textbf{\makecell[c]{Stochastic Loss}}} & \multicolumn{3}{c}{\textbf{\makecell[c]{Summation Loss}}} 
   % & \multicolumn{3}{c}{\textbf{GSM8K}} 
   \\ \midrule
   \multicolumn{7}{c}{\textbf{Auto-regressive Decoding}} \\
   \textbf{Model} & \textbf{Time per Token (ms)} & \textbf{Accuracy} & \textbf{Rejection Ratio} & \textbf{Time per Token (ms)} & \textbf{Accuracy} & \textbf{Rejection Ratio} 
   % & \textbf{Time per Token (ms)} & \textbf{Accuracy} & \textbf{Rejection Ratio}
   \\ 
   Layer 40 (full) & 96.41 & 23.95 & - & 86.10 & 25.24 & - 
   % & 93.60 & 33.05 & -
   \\
   \midrule
   \multicolumn{7}{c}{\textbf{Speculative Decoding}} \\
   \textbf{Draft Model} & \textbf{Time per Token (ms)} & \textbf{Accuracy} & \textbf{Rejection Ratio} & \textbf{Time per Token (ms)} & \textbf{Accuracy} & \textbf{Rejection Ratio}
   \\ %\midrule 
   Layer 12 & 58.86 ($1.63\times$) & 22.28 & 0.35 & 63.68 ($1.35\times$) & 16.52 & 0.40 
   \\
   Layer 16 & 60.25 ($1.60\times$) & 23.42 & 0.24 & 63.89 ($1.34\times$) & 20.92 & 0.30
   \\ 
   Layer 20 & 65.92 ($1.46\times$) & 25.09 & 0.16 & 69.61 ($1.23\times$) & 21.98 & 0.22 
   \\ \midrule
   \multicolumn{7}{c}{\textbf{Confidence-based Early-Exit \cite{varshney2023accelerating}}}\\
   \textbf{Model} & \textbf{Time per Token (ms)} & \textbf{Accuracy} & \textbf{Rejection Ratio} & \textbf{Time per Token (ms)} & \textbf{Accuracy} & \textbf{Rejection Ratio}
   \\ %\midrule 
   Layer 12:40 & 46.02 ($2.09\times$) & 20.69 & - & 50.89 ($1.69\times$) & 24.63 & - 
   \\ \midrule
   \multicolumn{7}{c}{\textbf{Sorted Self-Speculative Decoding (Ours)}}\\
   Layer 12:40 & 47.93 ($2.01\times$) & 23.27 & 0.06 & 65.19 ($1.32\times$) & 24.79 & 0.07\\
   Layer 12:24 & 48.66 ($1.98\times$) & 24.33 & 0.14 & 63.58 ($1.35\times$) & 25.17 & 0.19\\
   \bottomrule
 \end{tabular}}
 
 \label{tab:speculative}
\end{table*}

\subsection{Accelerating the inference of Decoder-based Large Language Models Using SortedNet}

% \todo[inline]{explain instanace aware and cite lite}
% \todo[inline]{create a figure to explain}

% \begin{figure*}[t!]
%     \centering
%     \begin{subfigure}[t]{0.5\columnwidth}
%         \centering
%         \includegraphics[height=1.2in]{Figs/lite.pdf}
%         \caption{Lorem ipsum}
%     \end{subfigure}%
%     ~ 
%     \begin{subfigure}[t]{0.5\columnwidth}
%         \centering
%         \includegraphics[height=1.2in]{Figs/lite.pdf}
%         \caption{Lorem ipsum, lorem ipsum,Lorem ipsum, lorem ipsum,Lorem ipsum}
%     \end{subfigure}
%     \caption{Caption place holder}
% \end{figure*}

\begin{figure}
    \centering
    \begin{subfigure}%[t]
        \centering
        \includegraphics[width=0.4\textwidth]{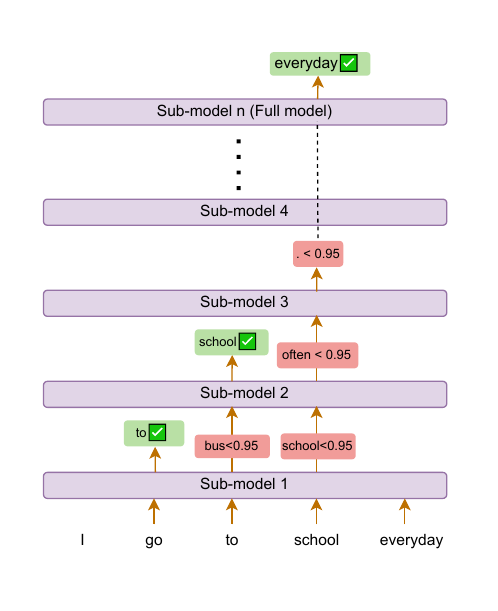}
        % \subcaption{(a)}
        \label{fig:bestsubset}
    \end{subfigure}%
    % \hfill
    \begin{subfigure}%[t]
        \centering
        \includegraphics[width=0.4\textwidth]{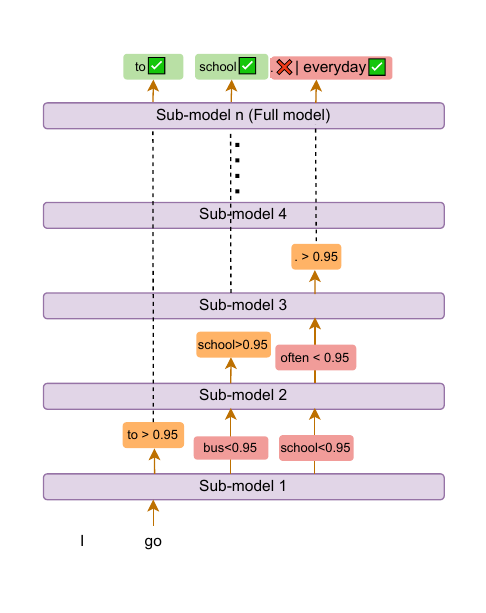}
        \label{fig:elbow}
        % \subcaption{(b)}
    \end{subfigure}
    \caption{(left) Confidence-based Early-Exit, exiting from intermediate sub-models whenever the confidence passes the determined threshold (right) Sorted Self-Speculative Decoding, verifying the adaptively exited draft tokens from intermediate sub-models by the full model.}
    \label{fig:sorted-spc}
\end{figure}

% \begin{figure}
%     \centering
%     \begin{subfigure}%[t]
%         \centering
%         \includegraphics[height=2.8in]{Figs/lite.pdf}
%         % \subcaption{(a)}
%         \label{fig:instance-aware}
%     \end{subfigure}%
%     \hfill
%     \begin{subfigure}%[t]
%         \centering
%         \includegraphics[height=2.8in]{Figs/lite.pdf}
%         \label{fig:cascade}
%         % \subcaption{(b)}
%     \end{subfigure}
%     \caption{(left) ? (right) ?}
% \end{figure}

% \begin{figure}[h!]
% \centering
% \resizebox{0.5\columnwidth}{!}{  
% \includegraphics[width=1.2\textwidth]{Figs/sorted-spc.pdf}
% }
% \caption{.}
% \label{fig:sortedspc}
% \end{figure}

To further show the scalability and generalizablity of SortedNet in more practical scenarios, we fine-tuned a LLaMA-13b \cite{touvron2023llama} on GSM8K \cite{cobbe2021training}, which is one of the challenging mathematical reasoning tasks. We chose the first 12, 16, 20, 24, 28, 32, 36, and 40 layers of LLaMA to build our submodels. To equalize the number of updates, we trained the model based on stochastic loss 8 times more than the summation loss, as we have 8 models and each forward pass in stochastic loss is 1/8 of the summation loss. In table \ref{tab:speculative}, we reported the performance of a subset of submodels and speedup gain that one can achieve using different sampling techinques such as Autoregressive decoding, speculative decoding \cite{leviathan2023fast}, Confidence-based Early-Exit which is a early-exiting approach based on the confidence of intermediate sub-models of the Sorted Models \cite{varshney2023accelerating}, and Sorted Self-Speculative Decoding, which adaptively utilizes intermeidate sub-models to generate draft tokens in Speculative Decoding algorithm (Figure \ref{fig:sorted-spc}). As shown, combining SortedNet and speculative decoding can improve the time per token efficiency up to 2.09 times faster than using auto-regressive for the full size model. In addition, we highlighted the details of each experiment hyperparameters in 
Appendix \ref{ap:hyperparameters} and further analysis has been provided in Appendix \ref{ap:sorting_impact} to better understand the behavior of sortedNet methodology. 

\section*{Conclusion}
In summary, this paper proposes a new approach for training dynamic neural networks that leverages the modularity of deep neural networks to efficiently switch between sub-models during inference. Our method sorts sub-models based on their computation/accuracy and trains them using an efficient updating scheme that randomly samples sub-models while accumulating gradients. The stochastic nature of our proposed method is helping our algorithm to generalize better and avoid greedy choices to robustly optimize many networks at once. We demonstrate through extensive experiments that our method outperforms previous dynamic training methods and yields more accurate sub-models across various architectures and tasks. 
The sorted architecture of the dynamic model proposed in this work aligns with sample efficient inference by allowing easier samples to exit the inference process at intermediate layers. Exploring this direction could be an interesting area for future work. %research.
% One other research direction which we will leave for future is to keep the gap between submodels of the SortedNet and the corresponding individual models trained regularly, as lowest as possible during each step. 

% \subsection*{Limitations}
% It is good to note that our proposed method might be sensitive to the randomness as the chosen trajectory at the moment is random uniform. Further research is necessary to investigate the effect of choosing more optimal strategies for choosing the next model at each iteration. One can further analyse the performance of progressively training submodels and ensure that the models converged properly at the ultimate training steps.

% \section*{References}
\bibliographystyle{unsrtnat}
\bibliography{custom}

\clearpage
\newpage

\appendix

\section*{Appendix}
\label{apdx:appendix}

\section{Related Work}
\label{ref::related}

\paragraph{ Slimmable Networks~\cite{yu_slimmable_2018}}
Slimmable networks is a method for training a single neural network in a way that it can be deployed with adjustable width at the inference time. This solution was proposed particularly for CNN architectures and thus, careful consideration of the batch normalization module for various width sizes is necessary. In this regard, in slimmable networks, switchable batch normalization was used which lead to additional trainable parameters. In contrast to slimmable networks, our SortedNet are architecture agnostic and work in both depth and width dimensions. 
% \moji{It's good to note that Slimmable Networks is the summation loss!}

\paragraph{Early Exit~\cite{xin2020deebert}} Early exit refers to a technique which adds a classifier to intermediate layers of an already trained neural network. While the parameters of the main model are frozen, the parameters of the classifiers are updated in a separate fine-tuning process. In this approach, each of the classifiers and their subsequent network can be treated as an independent sub-model. While this solution is relatively straightforward, the performance of the sub-models lags significantly behind that of the main model. Also dedicating a separate classification head to each sub-model can significantly increase the memory demand at inference. 

\paragraph{Dayna-BERT~\cite{hou2020dynabert}}

Dyna-BERT presents a dynamic compression method for pre-trained BERT models, enabling flexible adjustments in model size, both in depth and width, during inference. While the objective introduced in the DynaBERT paper shares some similarities with our approach, there are several key distinctions.
Firstly, in DynaBERT, only a few subsets of the model are functional, whereas our SortedNet does not rely on such an assumption.
Secondly, DynaBERT requires an already trained teacher model and utilizes knowledge distillation, whereas our technique operates independently of knowledge distillation. Thirdly, DynaBERT necessitates a search for an optimal sub-model, whereas our solution is inherently \textit{search-free}. Lastly, DynaBERT's applicability is dependent on the architecture, whereas our approach is architecture-agnostic.

\paragraph{Layer-drop~\cite{fan2019reducing}} 
Layer-drop is a structured dropout solution at the training time which allows layer pruning at the inference time. Similar to DynaBERT, this solution is applied to pre-trained language models; however, in contrast to DynaBERT, Layer-drop only targets the depth of neural networks and not their width. 
In Layer-drop, there is no fixed training pattern and any layer can be dropped with a certain probability, which is referred to as drop rate. At the inference time, the number of active layers can be adjusted by the drop-rates that are seen during the training time of that network (i.e. to achieve the best performance on any other drop-rate value, the network needs to be re-trained.). Layer-drop works only in depth while our solution works for both depth and width. Moreover, Layer-Drop requires specific search patterns for dropping layers at the inference time and training time, whereas our solution is search free. %does not have such a hyperparameter. It needs search at the training time for the layer dropout strategies and at the inference time to select the best layers.
 
\begin{figure}[htb!]
    \centering
    \includegraphics[width=0.45\textwidth]{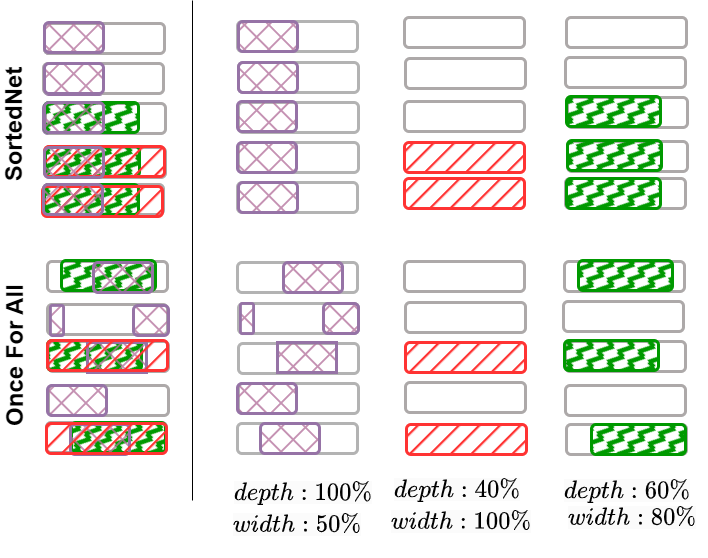}
    \caption{Comparing SortedNet and Once For All: on a hypothetical 5-layer network, we show how the sub-model selection strategy of SortedNet differs from the Once-for-All~\cite{cai_once-for-all_2020} approach.}
    \label{fig:comp}
\end{figure}

\paragraph{Once-for-All~\cite{cai_once-for-all_2020}}
Once-for-all(OFA) targets efficient inference across different devices by first training an OFA network which supports many sub-models with varying latency/accuracy characteristics ; it then searches among the feasible sub-models according to the accuracy and latency requirements of their target device. OFA has a progressive training nature i.e. it goes from the largest model to the smaller sub-models. OFA is different from our solution from the following aspects: first, it needs teacher and knowledge distillation; second, OFA requires a separate Neural Architecture Search (NAS) at the inference time; third, OFA is not architecture agnostic (their solution is for CNN-based neural networks while our SortedNet works for both CNNs and Transformers). Moreover, OFA is different from our solution in terms of the sub-model selection strategy. While our SortedNet selects sub-models in a sorted manner, OFA does not have any particular assumption for sorting sub-models (see Figure~\ref{fig:comp} for more details).   
% Also the arbitrary choice of procedural training will make it difficult to generalize this solution. 

\paragraph{Learning Compressible Subspace~\cite{Nunez_2023_WACV}}
Learning Compressible Subspace (LCS) is an adaptive compression technique based on training compressible subspace of neural networks (using a linear convex combination of two sets of weights for the network). While LCS does not require any re-training at the inference time,  this solution has some other limitations including: first, it needs double memory at the training time; second, the choices of initial weights and the compression function are unclear and arbitrary (left as a hyper-parameter); third, it is only tried on CNNs; forth, similar to Layer-drop intermediate sub-models are trained randomly which will make the performance of the target model sub-optimal.

\section{Theoretical Analysis} 
\label{ap:theory}
\subsection{Parameter Convergence in Identically Trained Sub-networks}

Suppose $\hat{f}$ is a sub-network within a larger neural network architecture, and $f$ represents an identical network architecture trained independently. We aim to understand the relationship between the parameters of these two networks, $\theta$ for $\hat{f}$ and $\phi$ for $f$, as they are trained under identical conditions.

\subsection{Assumption of Lipschitz Continuity of Gradients}
We assume that the gradients of the loss functions for $\hat{f}$ and $f$, denoted as $\mathcal{L}_{\hat{f}}$ and $\mathcal{L}_f$ respectively, are $L$-Lipschitz continuous. This implies that:
\begin{equation*}
    \|\nabla \mathcal{L}_{\hat{f}}(\theta) - \nabla \mathcal{L}_{\hat{f}}(\theta')\| \leq L \|\theta - \theta'\|
\end{equation*}
\begin{equation*}
    \|\nabla \mathcal{L}_f(\phi) - \nabla \mathcal{L}_f(\phi')\| \leq L \|\phi - \phi'\|
\end{equation*}
for all $\theta, \theta'$ and $\phi, \phi'$ in the parameter space.

\subsection{Parameter Update Rule}
The parameters of the networks are updated via gradient descent as follows:
\begin{itemize}
    \item For $\hat{f}$:
    \begin{equation*}
        \theta_{t+1} = \theta_t - \eta \nabla \mathcal{L}_{\hat{f}}(\theta_t)
    \end{equation*}
    \item For $f$:
    \begin{equation*}
        \phi_{t+1} = \phi_t - \eta \nabla \mathcal{L}_f(\phi_t)
    \end{equation*}
\end{itemize}

\subsection{Derivation of the Bound}
We derive a bound on the difference in parameters between $\hat{f}$ and $f$ after each training iteration:
\begin{equation*}
    \|\theta_{t+1} - \phi_{t+1}\| = \|\theta_t - \phi_t - \eta (\nabla \mathcal{L}_{\hat{f}}(\theta_t) - \nabla \mathcal{L}_f(\phi_t))\|
\end{equation*}
Applying the triangle inequality and the Lipschitz continuity of the gradients, we obtain:
\begin{equation*}
    \|\theta_{t+1} - \phi_{t+1}\| \leq (1 + \eta L) \|\theta_t - \phi_t\| + \eta C
\end{equation*}
where $C$ is a constant that bounds the difference between the gradients of the loss functions of the networks.

This bound quantifies the evolution of the difference in parameters between $\hat{f}$ and $f$ across training iterations, incorporating the impact of the Lipschitz constant $L$, the learning rate $\eta$, and the constant $C$ that bounds the inherent difference in gradients.

%%%%%%%%%%%%%

\subsection{Negligible C under Identical Training Conditions}
Given that $\hat{f}$ and $f$ are trained under perfectly identical conditions (same data, initialization, and hyperparameters), the difference in their gradients can be considered negligible, leading us to conclude that $C$ is practically zero. Under this assumption, the bound simplifies significantly:
\begin{equation*}
    \|\theta_{t+1} - \phi_{t+1}\| \leq (1 + \eta L) \|\theta_t - \phi_t\|
\end{equation*}
This indicates that the difference in the parameters of $\hat{f}$ and $f$ is governed by the Lipschitz constant $L$ and the learning rate $\eta$, suggesting that the parameters should remain close throughout the training process, especially when $C$ is negligible.

\subsection{Performance Bound} 
We would like to find a performance bound between a trained sub-model (with optimized parameters $\theta^*$) and its corresponding individual model (with optimized parameters $\phi^*$). 
% In this regard, supposed that we have a sub-model $\hat{f}(x;{\theta}^*)$ and its corresponding individual model ${f}(x;{\phi}^*)$. 
Let's assume that ${\phi}^* = {\theta^*} + \Delta \theta$.
Then, the performance bound can be calculated as $ \Delta {f} = {f}(x;{\phi^*}) - \hat{f}(x;{\theta^*}) $ in the function value from its optimal value due to a parameter perturbation

\textbf{Step 1: Second-Order Taylor Expansion}
Applying the second order taylor expansion to the function $f(x;\phi)$ around ($\phi = \phi^*$), we get: 

\begin{align*}
    & {f}(x; \phi^*) = {f}(x; \theta^* + \Delta \theta) \approx \\ 
    & {f}(x; \theta^*)  + \nabla_{\theta} {f}(x; \theta^*)^T \Delta \theta + \frac{1}{2} \Delta \theta^T H(x; \theta^*) \Delta \theta = \\ 
    & \hat{f}(x; \theta^*)  + \nabla_{\theta} \hat{f}(x; \theta^*)^T \Delta \theta + \frac{1}{2} \Delta \theta^T \hat{H}(x; \theta^*) \Delta \theta.
\end{align*}
Bear in mind that $f(x;\theta) = \hat{f}(x;\theta)$ and $H(x;\theta^*)$ refers to the Hessian matrix of $f$ at $(\theta = \theta^*)$.

\textbf{Step 2: Optimum Condition}
\begin{equation*}
    \nabla_{\theta} \hat{f}(x; \theta^*) = 0
\end{equation*}
\begin{equation*}
    \Rightarrow \hat{f}(x; \phi^*) \approx \hat{f}(x; \theta^*) + \frac{1}{2} \Delta \theta^T H(x; \theta^*) \Delta \theta
\end{equation*}

\textbf{Step 3: Lipschitz Continuity of Gradient}
\begin{equation*}
    \|\nabla_{\theta} \hat{f}(x; \theta) - \nabla_{\theta} \hat{f}(x; \theta')\| \leq L \|\theta - \theta'\|
\end{equation*}

\textbf{Step 4: Bounding the Hessian}
\begin{equation*}
    \|\Delta \theta^T \hat{H}(x; \theta) \Delta \theta\| \leq L \|\Delta \theta\|^2
\end{equation*}

\textbf{Step 5: Estimating the Deviation in Function Value}
\begin{align*}
    \Delta {f} &= {f}(x; \phi^*) - \hat{f}(x; \theta^*) \\
    \Delta {f} &\approx \frac{1}{2} \Delta \theta^T \hat{H}(x; \theta) \Delta \theta \leq \frac{1}{2} L \|\Delta \theta\|^2
\end{align*}

% \textbf{Conclusion}\\
The deviation \( \Delta {f} \) in the function value from its optimal value due to a parameter perturbation \( \Delta \theta \) is bounded by \( \frac{1}{2} L \|\Delta \theta\|^2 \) under the assumption of L-Lipschitz continuity of the gradient. This result implies that the function value's deviation grows at most quadratically with the size of the parameter perturbation.

\section{More Experimental Details}

\subsection{Ablation Study}

\paragraph{Convergence (Training Time) Analysis}
\label{convergence-time}

\begin{figure*}[htb!]
    \centering
    \resizebox{1.0\textwidth}{!}{  
     \begin{subfigure}%[b]{0.19\textwidth}
         \centering
         \includegraphics[width=1.0\textwidth]{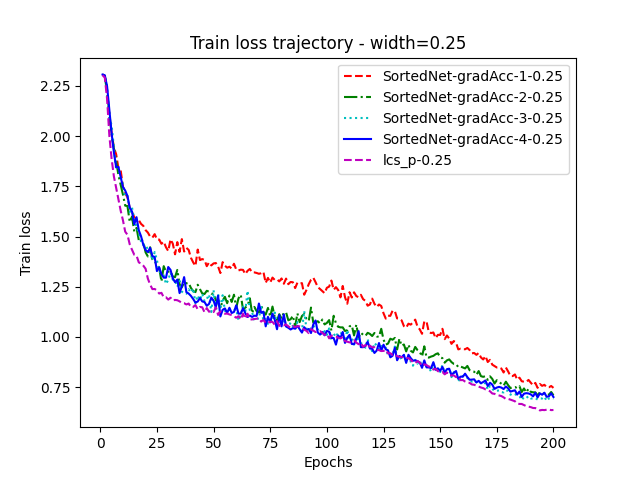}
         % \caption{a}
         \label{fig:accloss0.25}
     \end{subfigure}
     % \hfill
     \begin{subfigure}%[b]{0.19\textwidth}
         \centering
         \includegraphics[width=1.0\textwidth]{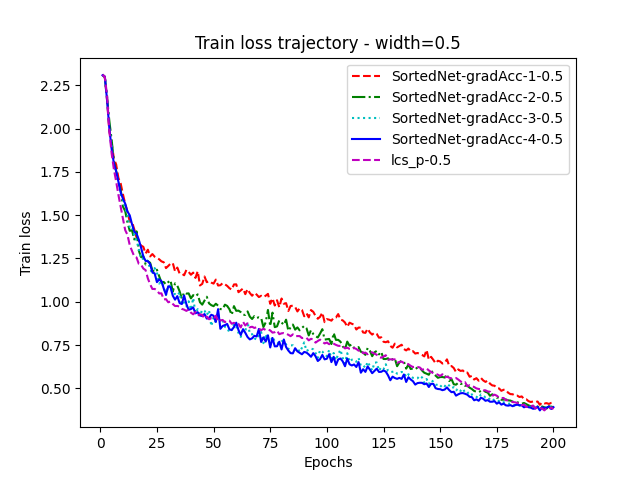}
         % \caption{}
         \label{fig:accloss0.5}
     \end{subfigure}
     % \hfill
     \begin{subfigure}%[b]{0.19\textwidth}
         \centering
         \includegraphics[width=1.0\textwidth]{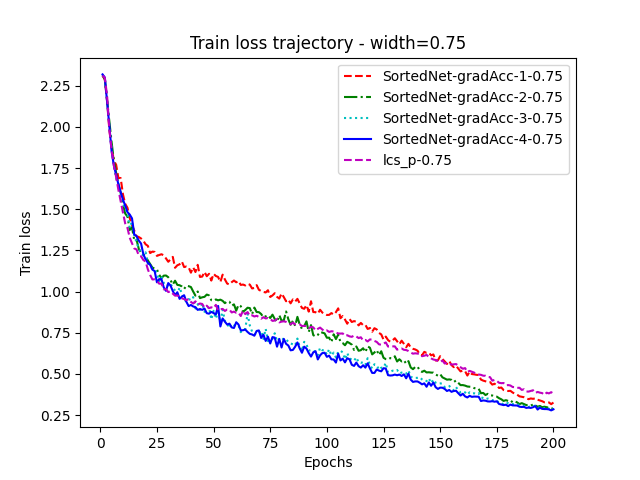}
         % \caption{}
         \label{fig:accloss0.75}
     \end{subfigure}
     \begin{subfigure}%[b]{0.19\textwidth}
         \centering
         \includegraphics[width=1.0\textwidth]{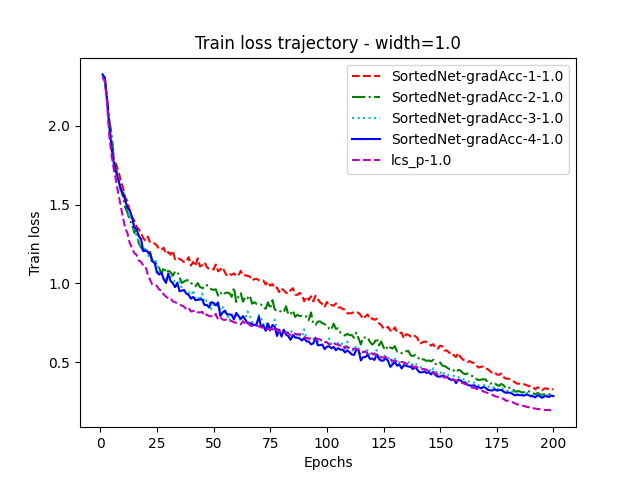}
         % \caption{}
         \label{fig:accloss1.0}
     \end{subfigure}
     \begin{subfigure}[b]%{0.19\textwidth}
         \centering
         \includegraphics[width=1.0\textwidth]{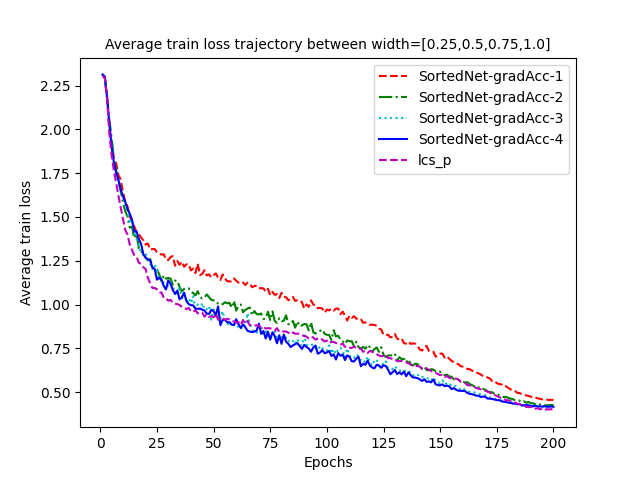}
         % \caption{}
         \label{fig:acclossavg}
     \end{subfigure}}
\caption{Comparing the training loss trajectory of SortedNet on CIFAR10 for different gradient accumulation values with LCS\_p. Each subfigure demonstrates the results in different widths. The rightmost subfigure reports the average across the widths. The underlying network (cPreResNet20) and hyperparameters are fixed.}
\label{fig:acc_loss_combined}
\end{figure*}

Being sorted and randomly selecting one sub-model at the time from a predefined set of the sub-models empowers SortedNet with a higher convergence rate and a faster training time. Figure \ref{fig:acc_loss_combined} empirically certifies this claim and compares the training convergence of SortedNet against LCP\_p, which, to the best of our knowledge, LCP\_p stands as the most recent state-of-the-art method. As LCS\_p uses summation loss over four sub-models in every training steps and to have a fair comparison, we therefore report the performance of SortedNet in different values of gradient accumulation ($g_{acc}$), where $g_{acc}=4$ provides a fair comparison with LCS\_p. As shown in the figure, SortedNet with $g_{acc}=4$ converges either faster or competitive across different sub-models. Moreover, SortedNet does not require any for-loop in its implementation; thus tailoring parallel computation and resulting in faster running time. We empirically investigate this feature and found that in the same setti

\paragraph{The impact of gradient accumulation}

The goal of this experiment is to examine the impact of gradient accumulation ($g_{acc}$) on the performance of SortedNet within an equal number of parameters updates. Table \ref{tab:gradientaccumulation} presents the results obtained in terms of accuracies for 4 different gradient accumulation values. To ensure an equal number of updates, the maximum number of epochs is adjusted for each scenario, e.g. $g_{acc}=k$ receives $k$ times more epochs than $g_{acc}=1$. As the results explains, increasing gradient accumulation values results in a higher performance for SortedNet. This observation can be attributed to the increase in training stochasticity when gradient accumulation is raised. Consequently, each sub-model in SortedNet contributes more equally to the updating of weight parameters, leading to a faster convergence rate. More details are provided in Appendix \ref{apdx:gradient-accumulation}. 

\subsection{Effect of gradient accumulation on SortedNet performance}
\label{apdx:gradient-accumulation}

\begin{table}[htb!]
\caption{Effect of gradient accumulation on SortedNet-IN performance while fixing the number of parameter updates. The underlying network and dataset are cPreResNet20 and CIFAR10, respectively. Numb. Updates refers to the number of calls to optimize.step()} 
% \vspace{5pt}
\centering
\resizebox{0.65\columnwidth}{!}{

% {|m{4.6cm}| x{2.6cm} c c x{1cm} x{2cm}  x{2cm} c x{3.2cm}| }

\begin{tabular}{lcccccccc}
\hline
Grad. Accum. & Num. Updates & Epochs & \multicolumn{4}{c}{Accuracy @ Width} & Avg.  \\
  &   &   & 100\%&75\%&50\%&25\% &  \\ 
% &aka optimizer.step() & & Width & Accuracy \\
\hline
 $g_{acc}=1$ &  200 & 200  & 84.94&84.92&82.54&71.03 &80.85  \\ 
 \hline
 $g_{acc}=2$ &  200 & 400  & 86.69&86.68&84.40&72.36 &82.53  \\
 \hline
 $g_{acc}=3$ &  200 & 600  & 87.37&87.50&84.57&73.00 &83.11  \\
 \hline
 $g_{acc}=4$ &  200 & 800  & 87.93&87.40&84.27&76.23 &83.95  \\
\hline

\end{tabular}}

\label{tab:gradientaccumulation}
\end{table}

It is of interest to explore whether limiting the number of parameter updates is a suitable approach for investigating the influence of gradient accumulation on SortedNet. One possible way to certify this factor is by running SortedNet with different gradient accumulation values while keeping the number of updates fixed. 
%For convenience, results have been reported in the same table \ref{tab:gradientaccumulation}.
To that end, we consider the same settings as Table \ref{tab:gradientaccumulation} and repeat the experiment while fixing the maximum number of training epochs. 
By fixing this value and increasing gradient accumulation values, we implicitly decrease the number of parameter updates. 
Table \ref{tab:gradientaccumulation-extra} reports the results. 
Comparing the results of these two tables, it is obvious that the number of updates plays a significant role in the model's performance. For instance, when considering $g_{acc}=2$, an average performance drop of approximately 2\% is observed across all sub-models. This reduction indicates that the underlying model needs more training time for higher values of $g_{acc}$.

\begin{table}[h!]
\caption{Exploring the impact of limited number of parameters updates on the effect of gradient accumulation in SortedNet-IN. The underlying network and dataset are cPreResNet20 and CIFAR10, respectively.}
\vspace{5pt}
\resizebox{\columnwidth}{!}{  
% \begin{tabular}{lcccc}
% \hline
% Gradient Acc. & Num. Updates & Epochs & \multicolumn{2}{c}{Extracted Networks} \\
% &aka optimizer.step() & & Width & Accuracy \\
% \hline
\begin{tabular}{lcccccccc}
\hline
Grad. Accum. & Num. Updates & Epochs & \multicolumn{4}{c}{Accuracy @ Width} & Avg.  \\
  &   &   & 100\%&75\%&50\%&25\% &  \\ 
 \hline
 $g_{acc}=1$ &  200 & 200  & 84.94&84.92&82.54&71.03 &80.85  \\ 
 \hline
$g_{acc}=2$ &  100 & 200  & 85.01&85.12&82.24&70.65 &80.75  \\ 
 \hline
 $g_{acc}=3$ &  66 & 200  & 85.09&85.06&82.64&73.74 &81.63  \\ 
 \hline
 $g_{acc}=4$ &  50 & 200  & 86.05&86.06&83.66&73.0 &82.19  \\ 
 \hline
\end{tabular}}

\label{tab:gradientaccumulation-extra}
\end{table}

\subsection{Hyperparameters}
\label{ap:hyperparameters}
This section provides an overview of the hyperparameters and experimental configurations, detailed in Table \ref{tab:hyperparamters-table}.

\begin{table}[htb!]
\caption{All the hyperparameters that have been used throughout our study for different experiments. If we didn't mention a parameter specifically, it means we utilized the default value of the HuggingFace Transformers v'4.27.0.dev0'\protect\footnote{https://huggingface.co/docs/transformers}. Otherwise, we highlighted any exception in the main text.}
\centering
\resizebox{0.7\columnwidth}{!}{  
\begin{tabular}{l|cc}
\hline
\textbf{Model} & \textbf{Parameter} & \textbf{Value} \\
\hline
\multirow{11}{*}{BERT-Base} & &  \\
& Optimizer & AdamW \\
& Warmup Ratio & 0.06 \\
& Dropout & 0.1 \\
& LR Scheduler & Linear \\
& Batch Size & 32 (RoBertA) / 8 (Bert) \\
& Epochs & 30 (RoBertA) / 3,6 (Bert) \\ 
& Learning Rate (LR) & 2e-5 (RoBertA / 6e-6 (Bert) \\
& Weight Decay & 0.1 \\
& Max Sequence Length & 512 \\
& Seeds & [10, 110, 1010, 42, 4242]\\
& GPU & Tesla V100-PCIE-32GB \\
\hline
\multirow{7}{*}{MobileNetV2} & &  \\
& Model & "google/mobilenet\_v2\_1.4\_224" \\
& Optimizer & AdamW \\
& LR Scheduler & Linear \\
& Batch Size & 128 \\
& Seeds & 4242\\
& Epochs & 60 $\times \#$ Models \\
& GPU & 8 $\times$ Tesla V100-PCIE-32GB \\
\hline
\multirow{9}{*}{cPreResNet20} & &  \\
& Optimizer & SGD \\
& Criterion & Cross Entropy \\
& LR Scheduler & cosine\_lr \\
& Batch Size & 128 \\
& Seeds & [40,42,1010,4242] \\
& Momentum & 0.9 \\
& Weight Decay & 0.0005 \\
& LR & 0.1 \\
& Epochs & [200,400,600,800] \\
& Gradient Accumulation & [1,2,3,4] \\

\hline
\end{tabular}}

\label{tab:hyperparamters-table}
\end{table}

\subsection{Details of training time comparison}
\label{apdx:running-time}
To empirically compare the training time between SortedNet and LCS\_p, the elapsed time per epoch for five epochs is recorded independently for each method. We then ignore the first epoch to reduce the impact of first-time loading and initialization. Next, for each method we take the average of the remaining elapsed times. We refer to these averaging times (in seconds) by $\bar{T}_{SortedNet}=49.7 \pm 2.06$ and $\bar{T}_{LCS\_p}=292.7 \pm 3.17$ for simplicity. As it is mentioned in Subsection \ref{convergence-time}, SortedNet with $g_{acc}=4$ can be considered as a fair comparison with LCS\_p. As a result, each epoch in LCS\_p holds four times the significance of SortedNet in terms of the total number of parameter updates. Therefore, we can simply multiply $\bar{T}_{SortedNet}$ by a factor of four to equalize their impacts in term of total number of parameter updates. By doing that, we have $\bar{T}_{SortedNet}=198.8$, which is almost one-third less than $\bar{T}_{LCS\_p}$.

\subsection{Can we improve the performance of SortedNet by adjusting the classifier layer?}
\label{ap:adjusting-classifier}

\begin{figure}[htb!]
\hspace*{-1cm}
\centering
\resizebox{0.8\columnwidth}{!}{  
    \includegraphics[]{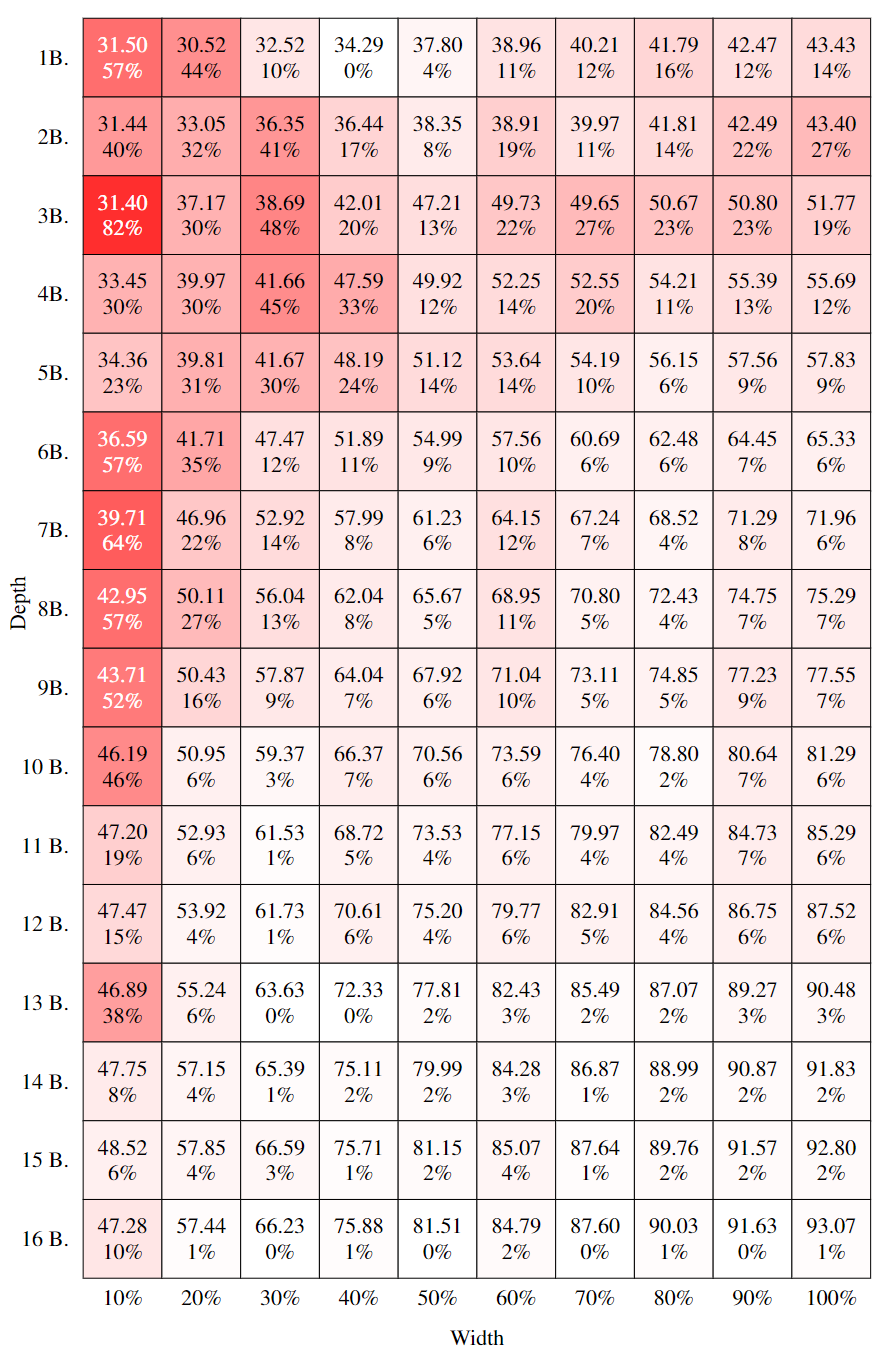}
}
\caption{CIFAR10 Adjusted Classification Accuracy for SortedNet (160 Models) and the baseline. The relative performance gain of each sub-model has been reported at the bottom of each cell with respect to the performance of the same network without adjustment. More white the better.}
\label{fig:scalable160-adjusted}
\end{figure}

In Figure \ref{fig:scalable160-bestsubset-left-elbow-right}-a, as mentioned before, we adjusted the performance of the classifiers for the baseline in the experiment but not for the SortedNet. Therefore, as an additional experiment, we wanted to analyze the impact of adjusting the classifier over the performance of our proposed method as well. Same as the previous experiment, we trained the classifier layer for 5 epochs for each sub-model and reported the performance. As shown in Figure \ref{fig:scalable160-adjusted}, the gain is much higher for very smaller networks than the large ones. The SortedNet shared classifier already doing a good job without additional computation overhead for all sub-models but further adjustments might be beneficial as shown.

\subsection{Can we extend SortedNet to complex dimensions?}
\label{ap:extend-more-complex}

\begin{table*}[h!]
\caption{The performance of BERT-base and Bert-large in the GLUE Benchmark over 5 runs for SortedNet (sharing weights across both models), pre-trained berts and different initialization. }
\centering
\resizebox{\textwidth}{!}{
\begin{tabular}{lccccccccccc}
\hline 
& & & Acc. & F1 & Acc. & Acc. & Matthews Corr. & Spearman Corr. & F1 & Acc. &  \\
Model & Flops & Weights & MNLI & QQP & QNLI & SST-2 & CoLA & STS-B & MRPC & RTE & avg. \\ 
\hline
\multicolumn{12}{c}{Random Initialized Networks} \\
\hline
$BERT_{BASE} (3\mathcal{L}_B)$ & 78.96 G  & $W_B$  & 62.13 $\pm$ 0.27 & 68.74 $\pm$ 0.70 & 61.24 $\pm$ 0.45 &  79.89 $\pm$ 0.59 & 0.00 $\pm$ 0.00 & 12.92 $\pm$ 0.63 & 78.67 $\pm$ 0.41 & 54.51 $\pm$ 1.11 & 52.26 \\ %$\pm$ 29.75 \\
\hline
\multicolumn{12}{c}{pre-trained Baselines} \\
\hline
$BERT_{BASE} (3\mathcal{L}_B)$* & 22.36 G & $W_B$ & 84.0 & 71.2 & 90.5 & 93.5 & 52.1 & 85.8 & 88.9 & 66.4 & 79.5 \\
$BERT_{LARGE} (3\mathcal{L}_L)$* & 78.96 G  & $W_L$ & 86.3 & 72.1 & 92.7 & 94.9 & 60.5 & 86.5 & 89.3 & 70.1 & 81.55 \\
\hline
\multicolumn{12}{c}{Paper Setting} \\
\hline
$BERT_{BASE} (3\mathcal{L}_B)$ & 22.36 G & $W_B$ & 84.22 $\pm$ 0.32 & 87.37 $\pm$ 0.08 & 91.47 $\pm$  0.21 & 92.61 $\pm$ 0.15 & 54.90 $\pm$ 0.79 & 88.08 $\pm$ 0.49 & 86.91 $\pm$ 0.82 & 62.96 $\pm$ 2.36 & \textbf{81.07} \\ %$\pm$ 14.08 \\
$BERT_{LARGE} (3\mathcal{L}_L)$ & 78.96 G  & $W_L$ & 86.32 $\pm$ 0.09 & 88.36 $\pm$ 0.07 & 92.01 $\pm$ 0.29 & 93.21 $\pm$ 0.42 & 59.39 $\pm$ 1.45 & 88.65 $\pm$ 0.33 & 88.67 $\pm$ 0.75 & 68.23 $\pm$ 1.59 & 83.11 \\ %$\pm$ 12.33 \\
\hline
\multicolumn{12}{c}{Extracted Networks} \\
\hline
$BERT_{BASE}^{LARGE} (3\mathcal{L}_B)$ & 22.36 G  & $W_B$ & 77.43 $\pm$ 0.08 & 84.88 $\pm$ 0.15 & 84.74 $\pm$ 0.34 & 84.98 $\pm$ 0.47 & 12.17 $\pm$ 1.62 & 78.33 $\pm$ 4.11 & 79.44 $\pm$ 0.93 & 55.23 $\pm$ 1.08 & 69.65 \\ %$\pm$ 25.18 \\
% $BERT_{BASE}^{LARGE} (6\mathcal{L}_B)$ & 22.36 G  & $W_B$ &  &  &  & & \\
\hline
\multicolumn{12}{c}{Proposed Methods} \\
\hline
Sorted $BERT_{BASE} (\sim1.5\mathcal{L}_B+1.5\mathcal{L}_L)$ & 22.36 G & $W_B^L$ & 76.20 $\pm$ 0.02 & 83.58 $\pm$ 0.16 & 83.91 $\pm$ 0.18 & 83.26 $\pm$ 0.69 & 0.08 $\pm$ 0.18 & 70.75 $\pm$ 9.25 & 80.75 $\pm$ 1.29 & 52.85 $\pm$ 2.53 & 66.42 \\ %$\pm$ 28.76 \\
Sorted $BERT_{LARGE} (\sim1.5\mathcal{L}_B+1.5\mathcal{L}_L)$ & 78.96 G & $W_L^L$ & 85.93 $\pm$ 0.33 & 87.28 $\pm$ 0.14 & 91.58 v 0.33 & 93.17 $\pm$ 0.26 & 57.08 $\pm$ 1.91 & 88.18 $\pm$ 0.68 & 87.06 $\pm$ 1.02 & 65.56 $\pm$ 1.41 & 81.98 \\ %$\pm$ 13.17 \\
Sorted $BERT_{BASE} (\sim3\mathcal{L}_B+3\mathcal{L}_L)$ & 22.36 G & $W_B^L$ & 77.48 & 85.16 $\pm$ 0.02 & 84.96 $\pm$ 0.23 & 86.01 $\pm$ 0.62 & 12.58 $\pm$ 2.04 & 79.29 $\pm$ 2.80 & 78.96 $\pm$ 0.44 & 55.81 $\pm$ 1.37 & 70.03 \\ %$\pm$ 25.16 \\
Sorted $BERT_{LARGE} (\sim3\mathcal{L}_B+3\mathcal{L}_L)$ & 78.96 G & $W_L^L$ & 86.12 & 88.26 $\pm$ 0.01 & 92.18 $\pm$ 0.28 & 93.49 $\pm$ 0.21 & 59.84 $\pm$ 1.35 & 88.85 $\pm$ 0.44 & 88.88 $\pm$ 1.10 & 68.45 $\pm$ 2.11 & \textbf{83.26}  \\ %$\pm$ 12.24 \\
\hline
\end{tabular}}

\label{tab:bert_base_large_ft}
\end{table*}

In this section, we are interested to investigate whether SortedNet is applicable to more complex dimensions other than width and depth. For example, can we utilize the SortedNet for sorting the Attention Heads \cite{vaswani2017attention}. To achieve this, we conducted an experiment over BERT-large \cite{devlin2019bert} which we tried to sort the information across multiple dimensions at once including, number of layers, hidden dimension, and number of attention heads. In other words, we tried to sort information over Bert-large and Bert-base as Bert-base can be seen as a subset of the Bert-large and therefore respect the nested property. As reported in Table \ref{tab:bert_base_large_ft}, in addition to the reported performance of Bert-base and Bert-large according to the original paper \cite{devlin2019bert}, we reported the performance of these models in the paper experimental setting. The performance of randomly initialized Bert-base has been reported as well. We also extracted a Bert-base from a Bert-large model, and we reported the performance of such model in the same table. Additionally, we highlighted the number of training updates with respect to each objective function in front of each model. For example, in the last row (Sorted $BERT_{LARGE}$), we approximately trained our Sorted model half of the times ($\sim3 Epochs$) over the objective function of Bert-base ($\mathcal{L}_B$) and the other half of the times over the objective function of Bert-large ($\mathcal{L}_L$) in an iterative random manner as introduced in the section \ref{sec:method}. The learned Bert-base performance with these methods is still around 10\% behind a pre-trained base but we argue that this is the value of pre-training. To investigate the impact, one should apply the SortedNet during pre-training which we will leave for future research. However, the performance of the learned Bert-large is on-par with an individual Bert-large which suggests sharing the weights does not necessarily have a negative impact over learning. It seems, however, the secret sauce to achieve a similar performance is that we should keep the number of updates for each objective the same as the individual training of Bert-large and Bert-base.

\subsection{What is the impact of Sorting?}
\label{ap:sorting_impact}

In order to better understand the impact of sorting information, we designed an experiment that compare the dependency order of all the neurons in a sorted network. To keep the experiment simple, we designed a one layer neural network with 10 (hidden dimension) $\times$ 2 (input dimension) neurons as the hidden layer and a classifier layer on top of that which map the hidden dimension to predict the probabilities of a 4 class problem. The task is to predict whether a 2d point belong to a specific class on a synthetic generated dataset. We trained both Sorted Network and the ordinary one for 10 epochs and optimize the networks using Adam optimizer \cite{kingma2017adam} with the learning rate of 0.01 and batch size of 16.

\begin{figure}[htb!]
% \hspace*{-1cm}
\centering
\resizebox{0.7\columnwidth}{!}{  
    \includegraphics[width=1.0\textwidth]{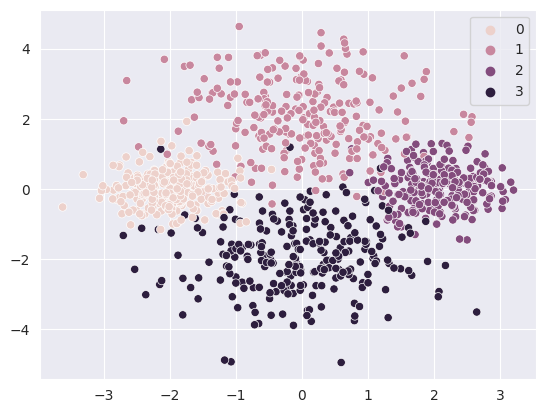}    
}
\caption{Synthetically generated dataset with four classes and with the centers of [[-2, 0], [0, 2], [2, 0], [0, -2]] and cluster standard deviation of [0.5, 1, 0.5, 1]. Seed has been fixed to 42, and 1000 samples has been generated.}
\label{fig:synthetic-dataset}
\end{figure}

\begin{table}[htb!]
\caption{Order dependency of all neurons in the network using the proposed method (SortedNet) and the ordinary training across 5 random runs. X means we used the neuron as is, and O means we removed the impact of that neuron by making it 0. $\uparrow$ higher the better, $\downarrow$ the lower the better.}
\centering
\resizebox{0.6\columnwidth}{!}{
\begin{tabular}{c|cc}
Active Neurons & Network Accuracy & SortedNet Accuracy \\
\hline
\multicolumn{3}{c}{Baseline} \\
\hline
\shortstack{XXXXXXXXXX \\ XXXXXXXXXX} &  \avercalc[2]{94.40, 94.50, 94.50, 94.30, 94.70} $\pm$ \stdcalc[2]{94.40, 94.50, 94.50, 94.30, 94.70} & \avercalc[2]{93.40, 93.20, 94.30, 93.70, 93.90} $\pm$ \stdcalc[2]{93.40, 93.20, 94.30, 93.70, 93.90}\\
\hline
 \shortstack{OOOOOOOOOO \\ OOOOOOOOOO} &  \avercalc[2]{25.0} $\pm$ \stdcalc[2]{25.0} & \avercalc[2]{25.0} $\pm$ \stdcalc[2]{25.0}\\
\hline
\multicolumn{3}{c}{Target Order $\uparrow$} \\
%%% 1
\hline
 \shortstack{XXXXXXXXXO \\ XXXXXXXXXO} &  \avercalc[2]{94.20, 93.60, 94.10, 93.40, 94.00} $\pm$ \stdcalc[2]{94.20, 93.60, 94.10, 93.40, 94.00} & \avercalc[2]{93.30, 93.00, 94.40, 93.60, 93.90} $\pm$ \stdcalc[2]{93.30, 93.00, 94.40, 93.60, 93.90}\\
%%% 2
\hline
 \shortstack{XXXXXXXXOO \\ XXXXXXXXOO} &  \avercalc[2]{93.60, 94.60, 94.30, 92.40, 93.60} $\pm$ \stdcalc[2]{93.60, 94.60, 94.30, 92.40, 93.60} & \avercalc[2]{93.30, 93.50, 94.30, 93.70, 94.10} $\pm$ \stdcalc[2]{93.30, 93.50, 94.30, 93.70, 94.10}\\
%%% 3
\hline
 \shortstack{XXXXXXXOOO \\ XXXXXXXOOO} &  \avercalc[2]{92.30, 94.0, 93.50, 93.50, 92.20} $\pm$ \stdcalc[2]{92.30, 94.0, 93.50, 93.50, 92.20} & \avercalc[2]{93.30, 92.80, 94.30, 93.70, 94.00} $\pm$ \stdcalc[2]{93.30, 92.80, 94.30, 93.70, 94.00}\\
 %%% 4
\hline
 \shortstack{XXXXXXOOOO \\ XXXXXXOOOO} &  \avercalc[2]{91.60, 92.60, 94.10, 92.10, 91.30} $\pm$ \stdcalc[2]{91.60, 92.60, 94.10, 92.10, 91.30} & \avercalc[2]{94.60, 92.30, 94.10, 93.70, 93.80} $\pm$ \stdcalc[2]{94.60, 92.30, 94.10, 93.70, 93.80}\\
 %%% 5
\hline
 \shortstack{XXXXXOOOOO \\ XXXXXOOOOO} &  \avercalc[2]{91.90, 91.90, 91.20, 89.70, 87.20} $\pm$ \stdcalc[2]{91.90, 91.90, 91.20, 89.70, 87.20}  & \avercalc[2]{94.50, 92.60, 94.10, 94.00, 94.10} $\pm$ \stdcalc[2]{94.50, 92.60, 94.10, 94.00, 94.10}\\
 %%% 6
\hline
 \shortstack{XXXXOOOOOO \\ XXXXOOOOOO} &  \avercalc[2]{92.80, 91.90, 85.50, 92.40, 71.60} $\pm$ \stdcalc[2]{92.80, 91.90, 85.50, 92.40, 71.60}  & \avercalc[2]{94.40, 92.10, 93.70, 94.00, 94.10} $\pm$ \stdcalc[2]{94.40, 92.10, 93.70, 94.00, 94.10}\\
 %%% 7
\hline
 \shortstack{XXXOOOOOOO \\ XXXOOOOOOO} &  \avercalc[2]{93.60, 82.40, 69.40, 89.20, 66.10} $\pm$ \stdcalc[2]{93.60, 82.40, 69.40, 89.20, 66.10}  & \avercalc[2]{94.10, 90.20, 94.50, 93.20, 93.70} $\pm$ \stdcalc[2]{94.10, 90.20, 94.50, 93.20, 93.70}\\
 %%% 8
\hline
 \shortstack{XXOOOOOOOO \\ XXOOOOOOOO} &  \avercalc[2]{50.60, 78.40, 43.00, 56.50, 76.30} $\pm$ \stdcalc[2]{50.60, 78.40, 43.00, 56.20, 76.30} & \avercalc[2]{93.50, 57.80, 91.40, 90.60, 93.00} $\pm$ \stdcalc[2]{93.50, 57.80, 91.40, 90.60, 93.00}\\
 %%% 9
\hline
 \shortstack{XOOOOOOOOO \\ XOOOOOOOOO} &  \avercalc[2]{49.20, 50.90, 46.30, 48.10, 48.80} $\pm$ \stdcalc[2]{49.20, 50.90, 46.30, 48.10, 48.80} & \avercalc[2]{48.30, 64.40, 62.10, 66.70, 71.40} $\pm$ \stdcalc[2]{48.30, 64.40, 62.10, 66.70, 71.40}\\
\hline
avg. & \avercalc[2]{93.86, 93.70, 93.10, 92.34, 90.38, 86.84, 80.14, 60.96, 48.66} $\pm$ \stdcalc[2]{93.86, 93.70, 93.10, 92.34, 90.38, 86.84, 80.14, 60.96, 48.66} & \textbf{\avercalc[2]{93.64, 93.78, 93.62, 93.70, 93.86, 93.66, 93.14, 85.26, 62.58}} $\pm$ \textbf{\stdcalc[2]{93.64, 93.78, 93.62, 93.70, 93.86, 93.66, 93.14, 85.26, 62.58}}\\
\hline
\multicolumn{3}{c}{Reverse Order $\downarrow$} \\
%%% 1
\hline
 \shortstack{OOOOOOOOOX \\ OOOOOOOOOX} &  \avercalc[2]{49.50, 58.80, 48.80, 48.20, 53.20} $\pm$ \stdcalc[2]{49.50, 58.80, 48.80, 48.20, 53.20} & \avercalc[2]{25.0, 50.50, 23.70, 25.00, 25.00} $\pm$ \stdcalc[2]{25.0, 50.50, 23.70, 25.00, 25.00}\\
 %%% 2
\hline
 \shortstack{OOOOOOOOXX \\ OOOOOOOOXX} &  \avercalc[2]{86.00, 92.00, 89.90, 66.60, 93.10} $\pm$ \stdcalc[2]{86.00, 92.00, 89.90, 66.60, 93.10} & \avercalc[2]{25.0, 85.30, 24.90, 15.70, 25.00} $\pm$ \stdcalc[2]{25.0, 85.30, 24.90, 15.70, 25.00}\\
 %%% 3 
\hline
 \shortstack{OOOOOOOXXX \\ OOOOOOOXXX} &  \avercalc[2]{91.90, 88.10, 93.20, 91.30, 89.60} $\pm$ \stdcalc[2]{91.90, 88.10, 93.20, 91.30, 89.60} & \avercalc[2]{25.0, 85.80, 33.50, 16.00, 47.30} $\pm$ \stdcalc[2]{25.0, 85.80, 33.50, 16.00, 47.30}\\
  %%% 4
\hline
 \shortstack{OOOOOOXXXX \\ OOOOOOXXXX} &  \avercalc[2]{89.20, 79.90, 86.40, 86.90, 91.10} $\pm$ \stdcalc[2]{89.20, 79.90, 86.40, 86.90, 91.10} & \avercalc[2]{48.40, 85.40, 61.50, 56.80, 47.30} $\pm$ \stdcalc[2]{48.40, 85.40, 61.50, 56.80, 47.30}\\
   %%% 5
\hline
 \shortstack{OOOOOXXXXX \\ OOOOOXXXXX} &  \avercalc[2]{93.10, 91.40, 93.50, 86.50, 90.10} $\pm$ \stdcalc[2]{93.10, 91.40, 93.50, 86.50, 90.10} & \avercalc[2]{48.50, 89.40, 89.50, 53.80, 49.10} $\pm$ \stdcalc[2]{48.50, 89.40, 89.50, 53.80, 49.10}\\
  %%% 6
\hline
 \shortstack{OOOOXXXXXX \\ OOOOXXXXXX} &  \avercalc[2]{93.60, 93.20, 93.30, 92.40, 92.20} $\pm$ \stdcalc[2]{93.60, 93.20, 93.30, 92.40, 92.20} & \avercalc[2]{48.40, 87.50, 90.80, 51.60, 49.30} $\pm$ \stdcalc[2]{48.40, 87.50, 90.80, 51.60, 49.30}\\
  %%% 7
\hline
\shortstack{OOOXXXXXXX \\ OOOXXXXXXX} &  \avercalc[2]{93.70, 93.70, 93.20, 93.80, 93.20} $\pm$ \stdcalc[2]{93.70, 93.70, 93.20, 93.80, 93.20} & \avercalc[2]{57.00, 90.30, 93.00, 68.80, 51.60} $\pm$ \stdcalc[2]{57.00, 90.30, 93.00, 68.80, 51.60}\\
   %%% 8
\hline
 \shortstack{OOXXXXXXXX \\ OOXXXXXXXX} &  \avercalc[2]{93.70, 94.00, 93.70, 93.30, 94.50} $\pm$ \stdcalc[2]{93.70, 94.00, 93.70, 93.30, 94.50} & \avercalc[2]{90.60, 91.50, 91.60, 68.10, 68.50} $\pm$ \stdcalc[2]{90.60, 91.50, 91.60, 68.10, 68.50}\\
    %%% 9
\hline
 \shortstack{OXXXXXXXXX \\ OXXXXXXXXX} &  \avercalc[2]{93.70, 94.00, 94.50, 94.20, 93.80} $\pm$ \stdcalc[2]{94.0, 94.00, 94.50, 94.20, 93.80} & \avercalc[2]{93.20, 92.80, 92.70, 62.60, 69.70} $\pm$ \stdcalc[2]{93.20, 92.80, 92.70, 62.60, 69.70}\\
\hline
avg. & \avercalc[2]{51.70, 85.52, 90.82, 86.70, 90.92, 92.94, 93.52, 93.84, 94.04} $\pm$ \stdcalc[2]{51.70, 85.52, 90.82, 86.70, 90.92, 92.94, 93.52, 93.84, 94.04} & \textbf{\avercalc[2]{29.84, 35.18, 41.52, 59.88, 66.06, 65.52, 72.14, 82.06, 82.2}} $\pm$ \textbf{\stdcalc[2]{29.84, 35.18, 41.52, 59.88, 66.06, 65.52, 72.14, 82.06, 82.2}}\\
\hline
\end{tabular}}

\label{tab:order_dep}
\end{table}

As can be seen in Table \ref{tab:order_dep}, the performance of different orders in the original neural network training paradigm can be different and unfortunately there is no specific pattern in it. Therefore, if one search the whole space of different orders (from neuron 1 to neuron n, from neuron n to neuron 1, or even select a subset of neurons by different strategies i.e. for the half size network activate every other neurons like XOXOXOXOXO.) might find better alternatives that work even better than the desirable target order. In this example, the reverse order in average perform better than the target order (86.67\% versus 82.22\%). However, with the proposed method, we can clearly see that the target order performance consistently is much better than the reverse order (89.25\% versus 59.38\%). This means, we have been able to enforce the desirable target order as we wanted using our proposed method. For example, neuron 2 is more dependent to neuron 1 in SortedNet in comparison with the ordinary training. In another example, the last 5 neurons are more dependent to the first 5 neurons than other way around. As shown, the performance of the first five neurons is 93.86\% while the performance of the last five neurons is only 66.06\% in SortedNet. In other words, the gain of adding the last five neurons is quite marginal and most probably prunnable, while the first 5 neurons contains most of the valuable information. %For example neuron 2 is more dependent to neuron 1 than neuron 1 to others. TODO: add several other examples that show this in the experiment
It is of interest to further investigate the dependency of neurons to one another and with other metrics which we will leave for future research.

\end{document}